\pdfoutput=1
\documentclass[11pt]{article}

\usepackage{ACL2025}

\usepackage{times}
\usepackage{latexsym}

\usepackage[T1]{fontenc}
\usepackage[utf8]{inputenc}

\usepackage{microtype}

\usepackage{inconsolata}
\usepackage{marvosym}
\usepackage{graphicx}
\usepackage{xcolor}
\usepackage{xspace}
\usepackage{multirow}
\usepackage{tabularx}
\usepackage{float}
\usepackage[ruled,vlined]{algorithm2e}
\usepackage{wrapfig}
\usepackage{lipsum} % for filler text
\usepackage{algpseudocode}
\usepackage{subcaption}
\usepackage{amsthm}
\usepackage{enumitem}
\usepackage{amsmath} 
\usepackage{booktabs}
\usepackage{amsfonts}       % blackboard math symbols
\usepackage{nicefrac} 
\usepackage{hyperref}

\usepackage{soul}
\usepackage{dsfont}
\usepackage{xspace}
\newcommand{\ours}{\textsc{A$^3$Tune}\xspace}
\newcommand{\moe}{\textsc{A$^3$MoE}\xspace}

\usepackage{colortbl}
\usepackage{makecell}
\usepackage{dsfont}
\usepackage{bbm}

\title{Focus on What Matters: Enhancing Medical Vision-Language Models with Automatic Attention Alignment Tuning}

\author{
Aofei Chang$^1$,
\textbf{Le Huang}$^2$,
\textbf{Alex James Boyd}$^2$,
\\
\textbf{Parminder Bhatia}$^2$,
\textbf{Taha Kass-Hout}$^2$,
\textbf{Cao Xiao}$^{2 *}$,
\textbf{Fenglong Ma}$^1$\thanks{~~Corresponding authors.}\\
$^1$Pennsylvania State University,
$^2$GE Healthcare,
\\
$^1$\{aofei, fenglong\}@psu.edu, \\ $^2$\{Lena.Huang, Alex.Boyd, Parminder.Bhatia, Taha.Kass-Hout, Cao.Xiao\}@gehealthcare.com
}

\begin{document}
\maketitle
\begin{abstract}
Medical Large Vision-Language Models (Med-LVLMs) often exhibit suboptimal attention distribution on visual inputs, leading to hallucinated or inaccurate outputs. Existing mitigation methods primarily rely on inference-time interventions, which are limited in attention adaptation or require additional supervision. To address this, we propose \ours, a novel fine-tuning framework for Automatic Attention Alignment Tuning. \ours leverages zero-shot weak labels from SAM, refines them into prompt-aware labels using BioMedCLIP, and then selectively modifies visually-critical attention heads to improve alignment while minimizing interference. Additionally, we introduce a \moe module, enabling adaptive parameter selection for attention tuning across diverse prompts and images. Extensive experiments on medical VQA and report generation benchmarks show that \ours outperforms state-of-the-art baselines, achieving enhanced attention distributions and performance in Med-LVLMs.\footnote{Source code is available at \url{https://github.com/Aofei-Chang/A3Tune}}

\end{abstract}

\section{Introduction}

\begin{figure*}[h!]
  \centering
  \includegraphics[width=0.85\linewidth]{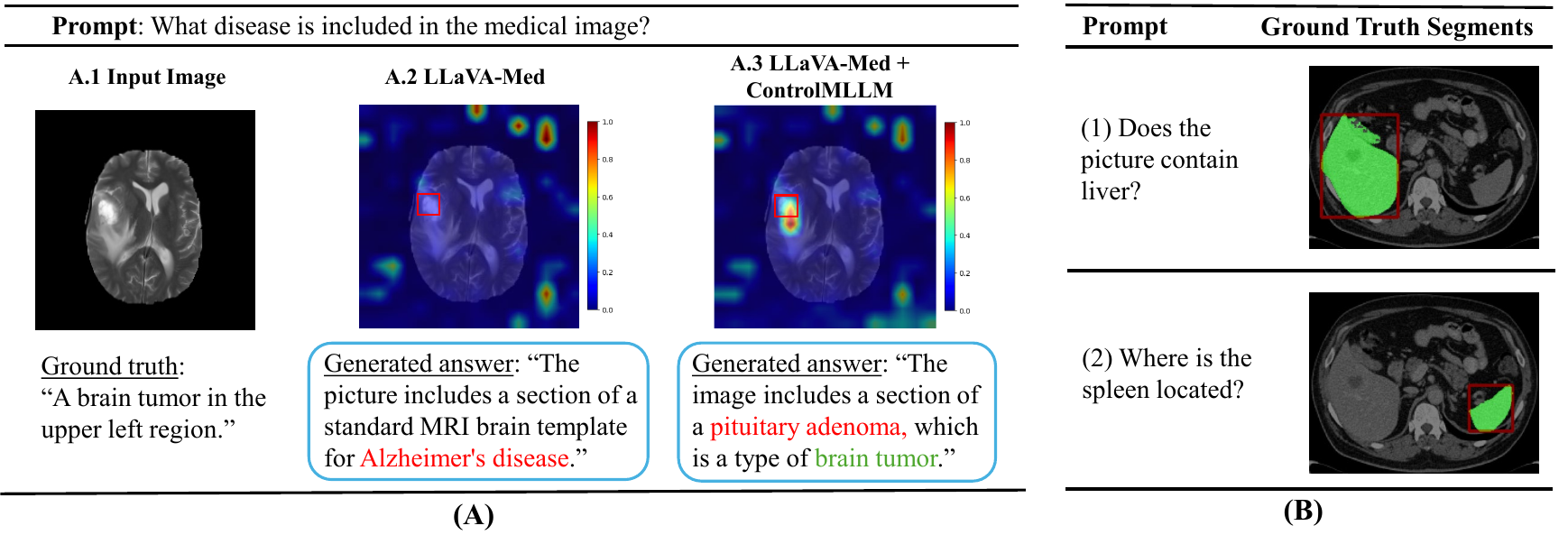}
  \vspace{-0.1in}
  \caption{(A) Examples of medical VQA and attention maps on medical images. In this example of Brain MRI from the SLAKE dataset, red box denotes the RoI of the brain tumor that LLaVA-Med should focus on. Red texts and green texts indicate wrong answers and correct answers, respectively. (B) Example of ground truth RoIs for different prompts on an Abdomen CT from SLAKE.}
  \vspace{-0.1in}
  \label{fig:preliminary}
\end{figure*}

While medical Large Vision-Language Models (Med-LVLMs) have shown significant progress in the medical domain~\cite{li2024llava, chexagent-2024, thawkar2023xraygpt, moor2023med}, they often produce inaccurate or hallucinated outputs that deviate from the provided visual medical information, as revealed by recent benchmarks~\cite{xia2024cares, gu2024medvh, chen2024detecting}. 
An example of medical visual question answering (VQA) extracted from the SLAKE~\cite{liu2021slake} dataset is shown in Figure~\ref{fig:preliminary}(A), where we visualize the average attention map on the image inputs during generating answers. We can observe that LLaVA-Med in Figure~\ref{fig:preliminary}(A.2) generates the hallucinated response ``\textit{Alzheimer's disease}'', neglecting the tumor region and over-focusing on irrelevant background areas, as shown in the corresponding attention map. This reveals a significant \textbf{bias in attention distribution} on visual inputs that limits the model’s effectiveness, which have been identified in general LVLMs~\cite{gong2024damro, woo2024don, liu2025paying}.

Unfortunately, \textit{none} of the specified bias mitigation strategies have been proposed in the medical domain. In the general domain, research primarily focuses on inference-time interventions to reduce attention biases, employing two main approaches.
The first approach, contrastive decoding~\cite{leng2024mitigating, favero2024multi, liu2025paying, woo2024don, gong2024damro}, introduces a contrastive adjustment to the decoding logits. However, this method does not directly modify the attention distribution, meaning it cannot guarantee that the model attends to diagnostically critical regions.
The second approach directly modifies attention maps during inference, as seen in ControlMLLM \cite{wu2024controlmllm}, which enforces the model to focus on pre-annotated regions of interest (RoIs). However, as shown in Figure\ref{fig:preliminary}(A.3), this method still generates partially hallucinated content. Besides, it requires additional tuning and ground truth RoIs for each inference process, making it impractical for real-world applications.

To overcome the limitations of inference-time intervention, an ideal solution is to automatically adjust attention maps towards RoIs during fine-tuning for downstream medical tasks. This approach enables the model to place more attention on critical regions during inference, eliminating the need for additional labels or interventions.  However, implementing such a method poses several challenges:

\textbf{(1) Limited availability of medical segmentation labels.} As shown in Figure~\ref{fig:preliminary}(A.3), using segmentation labels (RoIs) as guidance, as done in ControlMLLM, can enhance the learning of accurate attention maps and answer correctness. However, such labels are often unavailable in medical datasets.

\textbf{(2) Trade-off between attention alignment and model stability.} Assuming that RoI labels are available, directly modifying the attention maps of all attention heads towards the labels without any strategy is still risky. This may lead to the over-alignment issue that potentially impacts the output stability and overall performance. Therefore, achieving the right balance between attention alignment and model stability is essential.

\textbf{(3) Adapting attention to diverse prompts and images.} Even if attention alignment and model stability are balanced, the parameter-sharing strategy in fine-tuning remains a limitation for adaptive attention alignment. For example, as shown in Figure~\ref{fig:preliminary}, the optimal RoIs can vary significantly based on the input prompt and image, requiring dynamic alignment. While the computation of attention maps adjusts to input representations, the shared parameters in fine-tuning limit the model's ability to flexibly learn attention distribution across diverse inputs.

\begin{figure*}[t]
  \centering
  \includegraphics[width=0.85\linewidth]{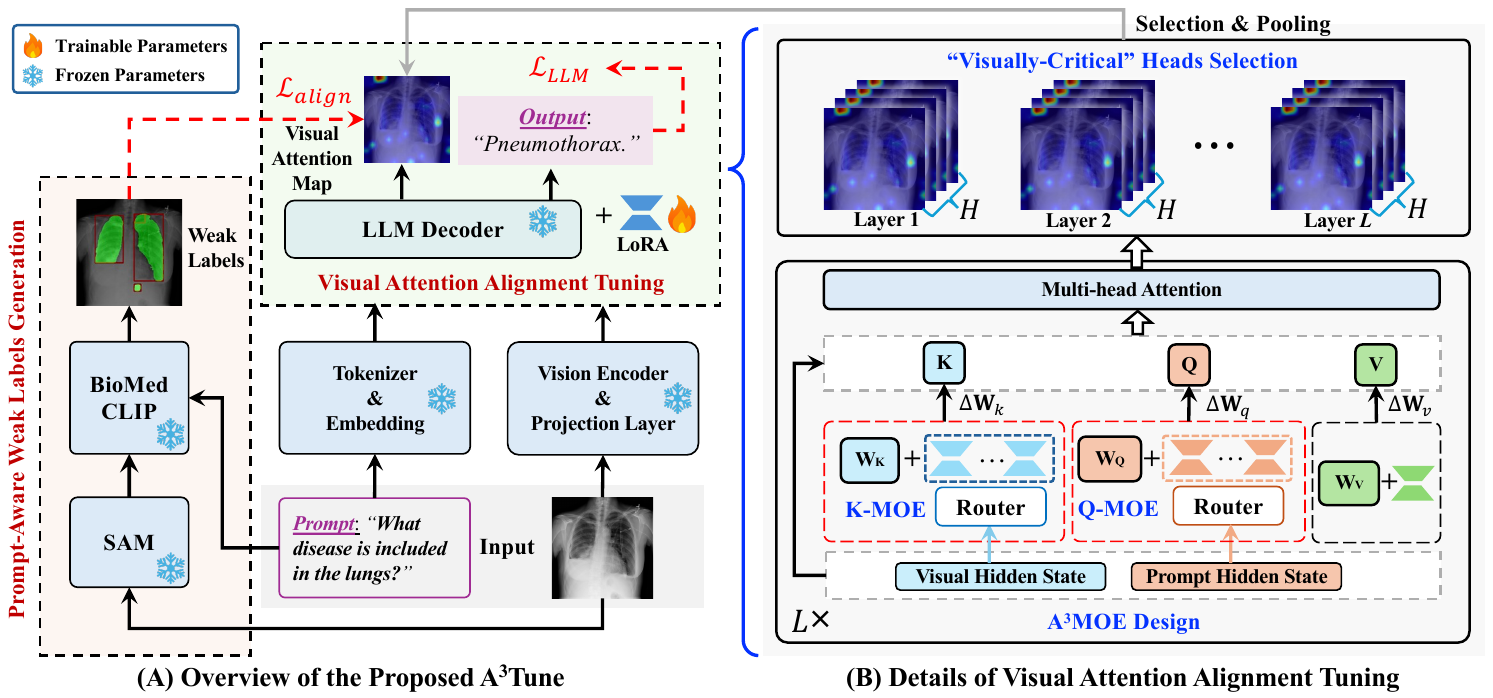}
  \vspace{-0.1in}
  \caption{(A) The overview of \ours and (B) the details of the designed visual attention alignment tuning.}
  \label{fig:model_design}
  \vspace{-0.1in}
\end{figure*}

To address these challenges simultaneously, we propose \ours, a novel fine-tuning framework designed for \textbf{A}utomatic \textbf{A}ttention \textbf{A}lignment \textbf{Tuning}. As shown in Figure~\ref{fig:model_design}, \ours integrates a  set of cooperative strategies to ensure that attention in Med-LVLMs is well aligned, minimally disruptive, and highly adaptable across diverse medical tasks. Firstly, to overcome the lack of segmentation labels (Challenge 1), \ours utilizes zero-shot segmentation labels generated by SAM~\cite{roy2023sam} and further refines them into prompt-aware weak labels using BioMedCLIP~\cite{zhang2023biomedclip}. These weak labels serve as guidance for attention alignment, eliminating the need for manual annotations. 
However, weak labels alone are not sufficient — uncontrolled modifications on all attention heads can disrupt model stability (Challenge 2). To further balance attention alignment with model stability, \ours selectively modifies only the most ``visually-critical'' attention heads, minimizing the risk of over-alignment and instability. Furthermore, the parameter-sharing strategy in alignment tuning remains a limitation, as RoIs vary significantly based on the input prompt and image (Challenge 3). To address this, we incorporate a custom-designed \textbf{M}ixture-\textbf{o}f-\textbf{E}xperts \moe into \ours on attention modules, allowing the model to dynamically select parameters and adjust attention maps for different images and prompts.

In summary, this work makes the following contributions:
(1) We propose \ours, a novel visual attention tuning approach that utilizes zero-shot weak labels to refine the visual focus of Med-LVLMs and enhance their performance.
(2) We develop a set of designs for cooperative attention alignment tuning: (i) weak label supervision, (ii) selective tuning of visually-critical attention heads, and (iii) a \moe module for adaptive attention adjustment.
(3) We conduct extensive experiments on five medical VQA and two report generation benchmarks against ten baselines, demonstrating that \ours outperforms state-of-the-art methods in both effectiveness and interpretability, improving visual grounding and overall model performance.

\section{Related Work}

% \subsection{Attention Biases in LVLMs}

% % (\ul{This paragraph can be moved to Related Work}) 
% Previous research has observed that attention often disproportionately focuses on meaningless background tokens, consuming a significant portion of the attention budget~\cite{gong2024damro} in LVLMs. This phenomenon can arise from the inherent properties of visual encoders like Vision Transformers, where global visual tokens are present, as analyzed in~\cite{darcet2024vision}. In the context of Med-LVLMs, such issues become even more pronounced, as observed in our analysis.
% \textbf{Hallucination Issues in Med-LVLMs.}
% Med-LVLMs, such as LLaVA-Med and LLaVA-Med-1.5, have demonstrated potential in medical applications but continue to suffer from significant factual inaccuracies. These models frequently produce responses that conflict with the visual information in medical images, as highlighted by recent benchmarks~\cite{xia2024cares, gu2024medvh, chen2024detecting}. While efforts have been made to evaluate and benchmark hallucinations, strategies for effectively mitigating these issues in Med-LVLMs remain underexplored.

% \noindent\textbf{Hallucination Mitigation.}
While some efforts, such as CoMT~\cite{Jiang2024CoMTCR}, have attempted to reduce hallucinations in Med-LVLMs for report generation by training on hierarchical QA pairs derived from real clinical image reports, mitigation strategies specifically designed for Med-LVLMs remain largely underexplored. Since Med-LVLMs share the same structure and training process as general LVLMs, hallucination issues are a common challenge across both. As a result, many inference-time mitigation strategies developed for LVLMs are also applicable to Med-LVLMs, including:
(1) Enhancing visual information and mitigating text over-reliance: VCD~\cite{leng2024mitigating}, M3ID~\cite{favero2024multi}, PAI~\cite{liu2025paying}, HELPD~\cite{yuan2024helpd}, RBD~\cite{liang2024mitigating}. (2) Mitigating visual token attention bias, with approaches such as AVISC~\cite{woo2024don} and DAMRO~\cite{gong2024damro}. (3) Refining and modifying LLM generation patterns during decoding, OPERA~\cite{huang2024opera}, DoLa~\cite{chuang2023dola}. In addition, other mitigation methods, such as TruthFlow~\cite{wang2025truthflow}, leverage trained steering vectors to guide the model toward more truthful outputs in LLMs. Despite these advancements, significant challenges remain in mitigation attention biases in Med-LVLMs. 
% (4) Post-hoc methods, including VOLCANO~\cite{lee2024volcano}, Woodpecker~\cite{yin2024woodpecker} and LURE~\cite{zhou2024analyzing}. 

% For our work, we adopt representative methods as baselines, excluding approaches that require additional targeted training such as post-hoc methods.

% While some methods, such as~\cite{Jiang2024CoMTCR}, have been proposed for medical report generation, they often require additional domain-specific data and targeted training. 

% \section{Preliminary Analysis and Motivations}
\section{Preliminaries}
\label{sec:preliminary}

% \begin{figure}[t]
%   \centering
%   \includegraphics[width=0.95\linewidth]{ACL2025-latex/figures/attention_map_placeholder.png}
%   \vspace{-0.1in}
%   \caption{Placeholder: The illustration of Visual Attention Matrix we focus on.}
%   \label{fig:attention_map_exp}
% \end{figure}

% \begin{figure}[t]
%   \centering
%   \includegraphics[width=0.95\linewidth]{ACL2025-latex/figures/preliminary_placeholder.png}
%   \vspace{-0.1in}
%   \caption{Placeholder: Cases in preliminary analysis.}
%   \label{fig:preliminary}
% \end{figure}

% \subsection{Background: LVLMs Modeling and Attention}
% \label{sec:bg_LVLM}
\subsection{Background of Med-LVLMs}
\label{sec:pre_architect}
Med-LVLMs share the same fundamental architecture as general LVLMs~\cite{liu2024visual, chen2023minigpt}, consisting of three primary components: the Visual Encoder, the Visual Alignment Layer, and the Large Language Model (LLM). The input image is firstly divided into \( N \) patches and processed by the Visual Encoder, which converts it into a sequence of visual tokens $\mathcal{T}_v$ with embeddings represented as  \( \mathbf{X}_v \in \mathbb{R}^{N \times d_v} \), where \( d_v \) denotes the dimension of the visual hidden representation. 

The visual tokens are then projected through the Visual Alignment Layer into \( \mathbf{X}_v^{'} \in \mathbb{R}^{N \times d_p} \), where \( d_p \) denotes the dimension of the LLM token space. These aligned visual tokens are subsequently forwarded to the LLM along with the embeddings \( \mathbf{X}_p \in \mathbb{R}^{V \times d_p} \) of the tokenized textual prompts $\mathcal{T}_p$ , where \( V \) is the number of text tokens. 

The final objective of the LLM in Med-LVLMs is to predict the next token \( y_t \) based on the current visual input $\mathbf{X}_v^{'}$ , prompt input $\mathbf{X}_p$, and previously generated tokens $y_{<t}$, formulated as:
\begin{equation} 
\label{eq:llm_modeling}
    p_t = p(y_t \mid \mathbf{X}_v^{'}, \mathbf{X}_p, y_{<t}; \Theta^{*}) \text{,}
\end{equation} where $\Theta^{*}$ denotes the parameters of the LLM.

\subsection{Visual Attention Map} 
\label{sec:pre_llm}
Most Med-LVLMs adopt a Transformer~\cite{Vaswani2017AttentionIA} decoder-based LLM, which processes inputs through $L$ decoder layers, each equipped with a multi-head attention module. In layer $l$, the module includes $H$ attention heads, where each head $h$ (with $1 \leq h\leq H$) computes attention separately using their corresponding attention map $\mathbf{M}_{lh}$. This attention mechanism models relationships between visual and textual tokens. 

To analyze how \textit{visual tokens} contribute to text generation, we focus on the attention scores between visual tokens and subsequent text tokens, referred to as \textit{visual attention map} $\mathbf{M}_{lh}^v$, a submatrix of the overall attention map $\mathbf{M}_{lh}$. To gain a global understanding of attention distribution on visual tokens, the averaged visual attention map $\mathbf{M}^v$ can be obtained by aggregating attention across all heads and layers~\cite{wu2024controlmllm}:
\begin{equation}
\label{eq:avg_attn}
    \mathbf{M}^v = \frac{1}{LH} \sum_{l=1}^{L} \sum_{h=1}^{H} \mathbf{M}_{lh}^v.
\end{equation}

\section{The Proposed \ours}
\label{sec:method}

% \subsection{Overview}
\ours aims to automatically align attention to enhance visual grounding and improve the performance of Med-LVLMs. The fine-tuning pipeline is illustrated in Figure~\ref{fig:model_design}. Given an input image $I$ and prompt $P$, we firstly generate a set of prompt-aware weak labels $\mathcal{S}$ using a zero-shot method (Section~\ref{sec:method_weak_label}).  To achieve effective attention alignment (Section~\ref{sec:method_vat}) with the guidance of $\mathcal{S}$, we first identify visually-critical attention heads, then integrate an \moe design into each decoder layer to enable flexible attention distribution learning.

% In the forward pass of Med-LVLMs, the image and prompt are processed and then forwarded through LLM decoder layers.

% ($\S$Section~\ref{sec:method_selection_heads}) to balance tuning effectiveness while minimizing instability. These selected heads play a crucial role in aligning attention to focus on relevant visual regions ($\S$Section~\ref{sec:method_attention_tuning_func}). Additionally, we integrate an \moe design into each decoder layer for flexible attention distribution learning ($\S$Section~\ref{sec:method_moe}). The \moe routes experts on the Query and Key matrices using prompt hidden states and visual hidden states in each decoder layer, enabling adaptive visual attention learning.

\subsection{Prompt-Aware Weak Labels Generation}
\label{sec:method_weak_label} 

Given a medical image $I$, we first use SAM~\cite{kirillov2023segment} to generate a set of candidate segments $\mathcal{S}^*$ following~\cite{yang2024fine}. However, using all segments in $\mathcal{S}^*$ for attention tuning introduces noise, as only a subset is relevant to each prompt. Therefore, it is necessary to select prompt-aware segments as weak labels to adaptively guide attention tuning. To filter prompt-aware weak labels, we embed each segment $s \in \mathcal{S}^*$ into a feature representation $\mathbf{E}_s$ using BioMedCLIP's vision encoder, while the text prompt $P$ is embedded into $\mathbf{E}_P$ using its text encoder. We then select an adaptive threshold $\tau_K$ to select $K$ segments that are most similar to the text prompt based on cosine similarity (Sim) in the embedding space:
\begin{equation} \label{eq:s}
\mathcal{S} = \{ s \in \mathcal{S}^* \mid \text{Sim}(\mathbf{E}_s, \mathbf{E}_P) \geq \tau_K \}.
\end{equation}

\subsection{Visual Attention Alignment Tuning}
\label{sec:method_vat}

The goal of visual attention alignment tuning is to align averaged visual attention map $\mathbf{M}^v$ with fixed weak labels $\mathcal{S}$ (i.e., masking regions) via Eq.~\eqref{eq:s} during fine-tuning. To achieve this goal, we first design a new parameter-efficient fine-tuning strategy and select ``visually-critical'' attention heads to the alignment of attention maps with $\mathcal{S}$.

\subsubsection{\moe Design}
\label{sec:method_moe}

\begin{figure}[t]
  \centering
  \includegraphics[width=0.95\linewidth]{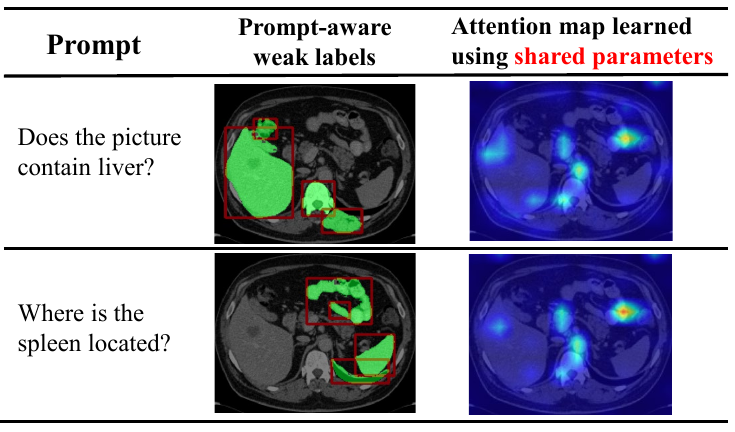}
  \vspace{-0.1in}
  \caption{Motivation for using \moe. The second column shows prompt-aware weak labels, with red bounding boxes and green inner segments. The third column shows the attention maps generated using shared parameters for the Query and Key matrices.}
  \label{fig:moe_motivation}
  \vspace{-0.15in}
\end{figure}

% To implement \ours, fine-tuning is required. However, directly tuning all parameters demands large scale medical data and high computation cost. To handle this, we 
We build \ours based on a parameter-efficient fine-tuning technique, LoRA~\cite{hu2021lora}, and apply it to all linear modules of the LLM in the Med-LVLM, including the Query and Key matrices in attention modules. 
In standard LoRA fine-tuning, the trainable LoRA parameters $\Delta \mathbf{W}_q$ and $\Delta \mathbf{W}_k$ for Query and Key matrices are shared across all training instances. However, this static parameter-sharing strategy in attention is insufficient for \ours. 
As shown in the abdominal organ analysis tasks (Figure~\ref{fig:moe_motivation}), weak labels $\mathcal{S}$ vary with the prompt, yet attention maps generated with shared $\Delta \mathbf{W}_q$ and $\Delta \mathbf{W}_k$ lack the flexibility to adapt effectively to different prompts and images. 

% Additionally, attention maps also need to vary across images, as each combination of image and prompt often requires tailored attention distributions to accurately capture fine-grained RoIs.

% $M^{v}$ in Equation~\ref{eq:l_attn} and it limits the model's ability to adapt attention distributions dynamically, which is critical for fine-grained, task-specific alignment.

% Intuitively, in medical tasks, the weak segmentation labels $S$ vary across images and prompts. Each combination of image and query often requires tailored attention distributions to effectively capture fine-grained RoIs. For example, different medical imaging modalities, such as chest X-rays and brain MRIs, demand distinct attention patterns based on the image and query. Relying on shared parameters limits the model's ability to specialize for these diverse inputs.

To address this limitation, we introduce \moe, a \textbf{M}ixture-\textbf{o}f-\textbf{E}xperts mechanism specifically designed for \ours, with two sub-modules: Q-MoE and K-MoE, applied to the LoRA parameters $\Delta \mathbf{W}_q^{(l)}$ and $\Delta \mathbf{W}_k^{(l)}$ in each LLM decoder layer $l$. For clarity, we omit the explicit annotation of $l$ in subsequent equations, as \moe is applied consistently across all decoder layers. The following sections describe these sub-modules in detail.

\noindent\ul{\textbf{Q-MoE: Prompt-Level MoE on Query Matrix.}} As shown in Figure~\ref{fig:moe_motivation}, the target RoIs often depend on the text prompt, even for the same image. To handle this, we introduce Q-MoE, a prompt-level MoE on $\Delta \mathbf{W}_q$, consisting of $O^q$ experts. For the $o$-th expert ($1 \leq o\leq O^q$),  we define $\mathcal{E}^q_o = \mathbf{B}_o \mathbf{A}_o$, where $\mathbf{B}_o \in \mathbb{R}^{ d_p \times r} $ and $\mathbf{A}_o \in \mathbb{R}^{ r \times d_p} $ are matrices with a low rank $r$. 

To dynamically route experts based on the prompt, we apply a prompt-level gating mechanism that generates router weights $\boldsymbol{\alpha}$ based on the hidden states of the prompt $\mathbf{H}_p$, which is obtained at each layer $l$ after processing $\mathbf{X}_p$ from the previous decoder layers. To capture the prompt's overall context, $\mathbf{H}_p$ is first averaged using a pooling operation. The routing and parameter computation are formulated as:
\begin{gather}
    \boldsymbol{\alpha} = \text{softmax}( \text{MLP} (\text{Pooling}(\mathbf{H}_p)) ), \\
    \Delta \mathbf{W}_q = \sum_{o=1}^{O^q} \boldsymbol{\alpha}_{o} \mathcal{E}^q_o ,
\end{gather}   

where $\boldsymbol{\alpha} \in \mathbb{R}^{O^q}$ represents the router weights, with each $\boldsymbol{\alpha}_{o}$ determining the contribution of the corresponding expert $\mathcal{E}^q_o$. MLP denotes the multi-layer perceptron used in the gating mechanism. 

\noindent\ul{\textbf{K-MoE: Visual Token-Level  Sparse MoE in Key Matrix.}}
Unlike text prompts, which can be summarized with pooling, visual inputs contain fine-grained information at the token level, where subtle differences can indicate abnormalities that require varied attention. To capture these nuances, we introduce K-MoE, a visual token-level MoE applied on $\Delta \mathbf{W}_k$. K-MoE includes $O^k$ experts, each implemented using LoRA. 

For each visual token $c$, the gating mechanism dynamically select experts based on its hidden states $\mathbf{H}_v^c$, and the LoRA parameters $\Delta \mathbf{W}^c_k$ for visual token $c$ are computed as:
\begin{gather}
    \boldsymbol{\beta}^c = \text{softmax}(\text{MLP} (\mathbf{H}_v^c)), \\
    \Delta \mathbf{W}^c_k = \sum_{o=1}^{O_k} \mathds{1}_{o}\boldsymbol{\beta}_{o}^{c} \mathcal{E}^k_o, 
\end{gather} 
where $\boldsymbol{\beta}^c \in \mathbb{R}^{O^k}$ and $\mathds{1}_{o}$ is a binary indicator enforcing sparsity by retaining only the top-$B$ gating weights. It is set to $1$ if expert $\mathcal{E}^k_o$ is among the top-$B$ most relevant experts, otherwise it is set to $0$. This mechanism ensures that for each visual token, only the most relevant experts contribute, minimizing interference among visual tokens while improving efficiency in alignment tuning.

% To ensure the alignment of attention maps with $\mathcal{S}$ while minimizing the intervention on fine-tuning and model output quality, we first identify ``visually-critical'' attention heads.

\subsubsection{``Visually-Critical'' Heads Selection}
\label{sec:method_selection_heads}
% We refine $M^v$ by selecting only the top-$k$ ``visually-critical'' heads. 

Each layer $l$ in the LLM decoder fine-tuned with \moe learns $H$ attention heads and the total number of attention heads is $L\times H$. Among them, the attention heads that assign higher weights to visual information are more crucial in processing visual features. Thus in layer $l$, the importance of each head $h$ can be quantified using the visual attention ratio $r_{lh}$, which measures the proportion of attention allocated to visual tokens relative to all tokens as follows:
\begin{equation}
    r_{lh} = \frac{\sum_{{c} \in \mathcal{T}_v} \mathbf{M}_{lh}^v[{c}]}{\sum_{c^{'} \in \mathcal{T}_v \cup \mathcal{T}_p} \mathbf{M}_{lh}^v[c^{'}]},
\end{equation}
where $\mathbf{M}_{lh}^v[c]$ represents the attention score assigned by attention head $h$ in layer $l$ to token $c$. 
% Higher $r_{lh}$ values indicate greater attention to visual tokens.

To refine the visual attention matrix $\mathbf{M}^v$, we select top-$R$ ``visually-critical'' heads based on the visual attention ratios $r_{lh}$. To achieve this, we use a binary indicator $\mathds{1}_{lh}$ to identify and retain only the most influential attention heads. Specifically, it is set to $1$ if attention head $(l, h)$ is among the $R$ heads with the highest visual attention ratios and $0$ otherwise.  The updated visual attention matrix is then computed as:
\begin{equation}
\label{eq:refine_attn_map}
    \widetilde{\mathbf{M}}^v = \frac{1}{R} \sum_{l=1}^{L} \sum_{h=1}^{H} \mathds{1}_{lh} \mathbf{M}_{lh}^v.
\end{equation}
% where $L$ and $H$ represent the number of layers and the number of heads in each layer. Selecting a smaller $R$ reduces interference to output quality but also weakens the effectiveness of attention alignment tuning.

\subsubsection{Attention Alignment Function}
\label{sec:method_attention_tuning_func}
After obtaining refined $\widetilde{\mathbf{M}}^v$, we define how it is adjusted to align with the weak labels $\mathcal{S}$. Intuitively, in the averaged attention map $\widetilde{\mathbf{M}}^v$, tokens within the masking regions $\mathcal{S}$ should receive higher attention values. Motivated by the cross-attention control design for image generation in~\cite{chen2024training}, we use a mask-based energy function to guide the attention alignment tuning:
\begin{equation}
\label{eq:l_attn}
\mathcal{L}_{\text{align}} = \sum_{s \in \mathcal{S}} \left( 1 - \frac{\sum_{c \in s} \widetilde{\mathbf{M}}_c^v}{\sum_{c^{'}=1}^N \widetilde{\mathbf{M}}_{c^{'}}^v} \right) ^2 ,
\end{equation}
where $s$ represents each referring segment in $\mathcal{S}$, $c$ and $c^{'}$ is the visual token index, and $N$ is the number of visual tokens. Among all $N$ visual tokens, this function encourages higher attention on visual tokens within each $s$, minimizing the loss and guiding $\widetilde{\mathbf{M}}^v$ to effectively focus on the prompt-aware regions $\mathcal{S}$.

% As discussed in Section~\ref{sec:pre_04}, rectifying the model's behavior can be effectively achieved by focusing on attention heads with high visual attention ratios. Therefore,

% By introducing these two sub-modules, \moe enables fine-grained, dynamic adaptation for diverse visual and text inputs in alignment tuning. 
% \begin{equation}
% \text{TopK}(v, k)_i = 
% \begin{cases} 
%   v_i & \text{if } v_i \text{ is among the top-$k$ largest values}\\ 
%   0 & \text{otherwise}
% \end{cases}
% \end{equation}

% For medical datasets, where inputs can range from different imaging modalities (e.g., X-rays, CT scans, MRIs) to varied diagnostic questions, such parameter sharing paradigm can result in suboptimal attention tuning results. Each combination of image and query often requires specialized attention distributions to capture critical regions effectively. Therefore, a more adaptable approach is necessary to enable attention tuning to dynamically adjust to each input.

\begin{table*}[t]
\centering

\caption{Performance comparison on medical VQA benchmarks using LLaVA-Med. The best results are highlighted in \textbf{bold}, and the second-best results are \ul{underlined}. OmniVQA corresponds to the OmniMedVQA dataset.}
\label{tab:main_exp_vqa}
\vspace{-0.1in}
\resizebox{1.85\columnwidth}{!}{
\begin{tabular}{c|c|cc|cc|cc|c|c}
\toprule
\multirow{2}{*}{\textbf{Model}} & \multirow{2}{*}{\textbf{Method}} & \multicolumn{2}{c|}{\textbf{Slake}} & \multicolumn{2}{c|}{\textbf{VQA-RAD}} & \multicolumn{2}{c|}{\textbf{PathVQA}} & \multicolumn{1}{c|}{\textbf{IU-Xray}} & \multicolumn{1}{c}{\textbf{OmniVQA}} \\ \cline{3-10}
 &  & \textbf{Open} & \textbf{Closed} & \textbf{Open} & \textbf{Closed} & \textbf{Open} & \textbf{Closed} & \textbf{Closed} & \textbf{Closed} \\
\midrule
% \multicolumn{9}{c}{\textit{Method based on LLaVA-Med Without Fine-tuning}} \\
\multirow{11}{*}{\textbf{LLaVA-Med}} 
& \cellcolor{gray!15}LLaVA-Med & \cellcolor{gray!15}41.28 & \cellcolor{gray!15}57.75 & \cellcolor{gray!15}32.48 & \cellcolor{gray!15}68.90 & \cellcolor{gray!15}10.34 & \cellcolor{gray!15}52.49 & \cellcolor{gray!15}72.83 & \cellcolor{gray!15}31.79 \\

& Greedy & 42.81 & 60.00 & 35.68 & 68.11 & 11.35 & 52.34 & 75.00 & 31.11  \\
& Beam~\cite{NIPS2014_beam} & 42.31 & 61.69 & 33.57 & 66.93 & 9.86 & 53.79 & 73.59 & 30.89  \\
& Nucleus~\cite{Holtzman2020The} & 41.06 & 61.41 & 32.64 & 68.11 & 9.93 & 53.17 & 73.09 & 30.96  \\
& VCD~\cite{leng2024mitigating} & 39.76 & 59.44 & 33.93 & 66.54 & 11.17 & 53.52 & 74.36 & 32.89 \\
& DoLa~\cite{chuang2023dola} & 42.37 & 59.72 & 35.68 & 68.50 & 11.38 & 52.55 & 74.87 & 31.04 \\
& OPERA~\cite{huang2024opera} & 40.44 & 60.00 & 35.54 & 68.50 & 9.77 & 53.17 & 75.00 & 31.72 \\
& AVISC~\cite{woo2024don} & 41.46 & 59.72 & 35.43 & 64.17 & 11.03 & 52.20 & 73.85 & 32.48 \\
& M3ID~\cite{favero2024multi} & 38.85 & 60.28 & 35.31 & 62.60 & 9.80 & 52.93 & 72.32 & 31.11 \\
& DAMRO~\cite{gong2024damro} & 41.33 & 59.72 & 32.73 & 66.54 & 10.80 & 51.84 & 72.19 & 31.45  \\
& PAI~\cite{liu2025paying} & 43.18 & 60.28 & 35.10 & 68.50 & 11.09 & 52.46 &  74.87 & 32.13 \\

\midrule
% \multicolumn{9}{c}{\textit{Method based on LLaVA-Med With LoRA Fine-tuning}} \\
% \midrule

\multirow{12}{*}{\textbf{\makecell{LLaVA-Med \\ + LoRA}}} 
& \cellcolor{gray!15}LLaVA-Med + LoRA & \cellcolor{gray!15}80.65 & \cellcolor{gray!15}82.82 & \cellcolor{gray!15}33.37 & \cellcolor{gray!15}66.54 & \cellcolor{gray!15}31.92 & \cellcolor{gray!15}90.95 & \cellcolor{gray!15}83.29 & \cellcolor{gray!15}90.65 \\

% & Greedy +  & 81.12  & 85.07 & 31.88 & 68.50 & 33.90 & 91.86 & 83.93 & 90.73  \\
% & Beam + & 81.32 & \textbf{86.76} & 32.55 & 68.90 & 33.60 & \ul{91.92} & 84.06 & 90.58\\
% & Nucleus + & 80.18 & 85.35 & 31.34 & 68.50 & 30.14 & 90.92 & 83.41& 90.05 \\
% & VCD + & 79.58 & 84.23 & 32.96 & 67.72 & 30.20 & 90.86 & 83.93&90.73\\
% & DoLa + & 81.84 & \ul{86.48} & 31.94 & 68.50 & \ul{34.00} & 91.86 & 84.06 &90.69 \\
% & OPERA + & 81.25 & \ul{86.48} & 33.18 & 68.90 & 33.64 & 91.80 & 84.06 &90.42 \\
% & AVISC + & 80.15 & 85.63 & 33.66 & \ul{69.29} & 32.37 & 90.62 & 84.06 & 90.54 \\
% & M3ID + & 79.83 & 84.79 & 31.40 & 68.90 & 32.47 & 91.15 & \ul{84.95} & 90.73\\
% & DAMRO + & \ul{82.19} & 83.66 & 32.41 & 66.14  & 32.27 & 90.12 & 84.69 & 90.08  \\
% & PAI + & 81.02 & \textbf{86.76} & 32.14 & 68.11 & 33.46 & 91.77 & 84.31 & \ul{90.95}\\

& Greedy  & 81.12  & 85.07 & 31.88 & 68.50 & 33.90 & 91.86 & 83.93 & 90.73  \\
& Beam~\cite{NIPS2014_beam} & 81.32 & \textbf{86.76} & 32.55 & 68.90 & 33.60 & \ul{91.92} & 84.06 & 90.58\\
& Nucleus~\cite{Holtzman2020The} & 80.18 & 85.35 & 31.34 & 68.50 & 30.14 & 90.92 & 83.41& 90.05 \\
& VCD~\cite{leng2024mitigating} & 79.58 & 84.23 & 32.96 & 67.72 & 30.20 & 90.86 & 83.93&90.73\\
& DoLa~\cite{chuang2023dola} & 81.84 & \ul{86.48} & 31.94 & 68.50 & \ul{34.00} & 91.86 & 84.06 &90.69 \\
& OPERA~\cite{huang2024opera} & 81.25 & \ul{86.48} & 33.18 & 68.90 & 33.64 & 91.80 & 84.06 &90.42 \\
& AVISC~\cite{woo2024don} & 80.15 & 85.63 & \ul{33.66} & \ul{69.29} & 32.37 & 90.62 & 84.06 & 90.54 \\
& M3ID~\cite{favero2024multi} & 79.83 & 84.79 & 31.40 & 68.90 & 32.47 & 91.15 & \ul{84.95} & 90.73\\
& DAMRO~\cite{gong2024damro} & \ul{82.19} & 83.66 & 32.41 & 66.14  & 32.27 & 90.12 & 84.69 & 90.08  \\
& PAI~\cite{liu2025paying} & 81.02 & \textbf{86.76} & 32.14 & 68.11 & 33.46 & 91.77 & 84.31 & \ul{90.95}\\
% \midrule
% \cline{2-10}
& \ours(ours) & \textbf{82.36} & \textbf{86.76} & \textbf{36.97} & \textbf{70.87} & \textbf{34.61} & \textbf{92.19} & \textbf{85.97} & \textbf{91.98}\\
\bottomrule
\end{tabular}
}
% \vspace{-0.1in}
\end{table*}

% rotate the columns and rows

\subsection{Final Objective}

In the fine-tuning process, we incorporate the loss of proposed visual attention alignment tuning $\mathcal{L}_{\text{align}}$ (Eq.~\ref{eq:l_attn})  as a regularization term for downstream medical tasks. This term is combined with the language modeling objective $\mathcal{L}_{LLM}$:
\begin{equation}
    \mathcal{L}_{LLM}  = -\sum_{t=1}^T \log p_t(y_t \mid \mathbf{X}_v^{'}, \mathbf{X}_p, y_{<t}; \Theta, \Theta^*),
\end{equation}
where $\Theta^*$ denotes the frozen parameters of LLM, $\Theta$ denotes the trainable parameters in LoRA and \moe, and $y_{<t}$ denotes the generated tokens before time step $t$. The final objective loss is:
\begin{equation}\label{eq:lambda}
    \mathcal{L} = \mathcal{L}_{LLM} + \lambda \mathcal{L}_{\text{align}},
\end{equation}
where $\lambda > 0$ is a hyperparameter that controls the strength of attention tuning.

\section{Experiments}

We evaluate our method on representative Med-LVLMs, including LLaVA-Med~\cite{li2024llava} and LLaVA-Med-1.5. 
The experimental results for LLaVA-Med-1.5 are provided in Appendix~\ref{appd:exp_llavamed1.5}. 
% The default values for key hyperparameters are: $K=4$, $R=128$, $B=3$. The value of $\lambda$ varies with the total fine-tuning steps for different datasets to dynamically control the alignment tuning strength. More implementation details can be found in Appendix~\ref{appd:implementation}.

\subsection{Settings}
% \noindent\ul{\textbf{Med-LVLMs and Hyperparameters.}} We evaluate our method on representative Med-LVLMs, including LLaVA-Med~\cite{li2024llava} and LLaVA-Med-1.5, to validate its effectiveness and generalizability. Experimental results for LLaVA-Med-1.5 are provided in Appendix~\ref{appd:exp_llavamed1.5}. The default values for key hyperparameters are: $K=4$, $R=128$, $B=3$. The value of $\lambda$ varies with the total fine-tuning steps for different datasets to dynamically control the alignment tuning strength. More implementation details can be found in Appendix~\ref{appd:implementation}.
% We evaluate our method on representative Med-LVLMs, including LLaVA-Med~\cite{li2024llava}, LLaVA-Med 1.5, and XrayGPT~\cite{thawkar2023xraygpt}, to validate its effectiveness and generalizability. The experimental results for LLaVA-Med 1.5 and XrayGPT are provided in Appendix~\ref{appd:exp_llavamed1.5} and Appendix~\ref{appd:exp_xraygpt}, respectively.

\noindent\ul{\textbf{Datasets.}}
We evaluate our method on two key tasks in medical application of Med-LVLMs: medical VQA and medical report generation. For medical VQA, we use diverse datasets including SLAKE~\cite{liu2021slake}, VQA-RAD~\cite{lau2018dataset}, PathVQA~\cite{he2020pathvqa}, IU-Xray~\cite{demner2016preparing} and OmniMedVQA~\cite{hu2024omnimedvqa}. For medical report generation, we use MIMIC-CXR~\cite{johnson2019mimic} and IU-Xray. Details of dataset processing and settings are provided in Appendix~\ref{appd:dataset}.

\noindent\ul{\textbf{Baselines.}}
We compare our approach with widely used hallucination mitigation methods with or without fine-tuning, including decoding strategies and  contrastive decoding techniques\footnote{We do not compare with the methods requiring tailored de-hallucination training pipelines such as CoMT~\cite{Jiang2024CoMTCR} and HELPD~\cite{yuan2024helpd}. }. The \textbf{decoding} baselines include Greedy decoding, Nucleus sampling~\cite{Holtzman2020The}, Beam search~\cite{NIPS2014_beam}. For \textbf{contrastive decoding} techniques, we specifically compare with: VCD~\cite{leng2024mitigating}, OPERA~\cite{huang2024opera}, DoLa~\cite{chuang2023dola}, 
AVISC~\cite{woo2024don}, M3ID~\cite{favero2024multi}, DAMRO~\cite{gong2024damro} and
PAI~\cite{liu2025paying}. Additionally, we include ControlMLLM~\cite{wu2024controlmllm} as a baseline only when the ground truth RoIs are available.
% Notably, the selected baselines apply inference-time interventions and are originally designed for models without fine-tuning. Since \ours is built on the fine-tuning process for downstream tasks, we ensure fair comparisons by evaluating baselines in both settings: (i) applied directly to the base model and (ii) applied to the fine-tuned model. 
Detailed settings of baselines and implementation are in Appendix~\ref{appd:implementation_baselines} and Appendix~\ref{appd:implementation}, respectively.

\begin{table}[t]
\centering
% \vspace{-0.1in}
\caption{Comparison of Visual Attention Distribution.}
\label{tab:interpret}
\vspace{-0.1in}
\resizebox{0.85\columnwidth}{!}{
\begin{tabular}{c|cc}
\toprule
\multirow{2}{*}{\textbf{Method}} & \multicolumn{2}{c}{\textbf{Metrics}} \\
 \cline{2-3}
& \textbf{Coverage $\uparrow$ } & \textbf{Intensity $\uparrow$} \\
\midrule
LLaVA-Med & 0.122 & 0.076 \\
LLaVA-Med + LoRA & 0.132 & 0.076 \\
\ours  & \textbf{0.275} & \textbf{0.147} \\
\bottomrule
\end{tabular}}
\vspace{-0.1in}
\end{table}

\begin{table*}[t]
\centering
\caption{Performance on report generation benchmarks using LLaVA-Med fine-tuned with LoRA.}
\label{tab:main_exp_report_02}
\vspace{-0.1in}
\resizebox{2\columnwidth}{!}{
\begin{tabular}{p{1.3cm}|p{1.8cm}|ccccccccccc|c}
\toprule
\multirow{2}{*}{\textbf{Dataset}} & \multirow{2}{*}{\textbf{Metric}} & \multicolumn{12}{c}{\textbf{Method}}\\
\cline{3-14}
& & \cellcolor{gray!15}\makecell{LLaVA-Med \\ + LoRA} 
& Greedy 
& Beam
& Nucleus
& VCD
& DoLa
& OPERA
& AVISC
& M3ID
& DAMRO
& PAI
% & \makecell{\ours \\ (ours)} \\
& \ours \\
\midrule
\multirow{7}{*}{\textbf{IU-Xray}} & \textbf{BLEU} & \cellcolor{gray!15}7.70 & 8.86 & \ul{9.34} & 7.80 & 8.83 & 8.93 & 8.27 & 8.52 & 8.63 & 7.29 & 8.87 & \textbf{11.05} \\
& \textbf{ROUGE-L}  & \cellcolor{gray!15}26.15 & 27.09 & 27.56 & 26.72 & 27.36 & 26.94 & 27.14 & 26.97 & \ul{27.79} & 25.61 & 26.74 & \textbf{30.00} \\
& \textbf{METEOR}  & \cellcolor{gray!15}29.50 & 26.01 & 26.44 & 30.33 & \ul{31.77} & 25.74 & 29.66 & 31.14 & 31.65 & 30.03 & 25.99 & \textbf{34.26} \\
& \textbf{BERTScore}  & \cellcolor{gray!15}88.36 & 88.50 & \ul{88.52} & 88.28 & 88.30 & 88.42 & 88.41 & 88.47 & \ul{88.52} &88.14 & 88.39 & \textbf{89.05} \\
& \textbf{CheXbert}  & \cellcolor{gray!15}53.81 & 52.55 & 52.88 & 52.73 & 51.86 & 52.27 & \ul{56.26} & 53.33 & 54.45 & 51.50 & 52.45 & \textbf{57.19} \\
& \textbf{RadGraph}  & \cellcolor{gray!15}20.44 & 20.76 & 21.29 & 20.85 & 22.02 & 20.63 & 21.87 & \ul{22.27} & 22.22 & 20.75 & 20.48 & \textbf{24.24} \\
& \textbf{RaTEScore}  & \cellcolor{gray!15}58.37 & 58.24 & 58.77 & 57.84 & 58.93 & 58.10 & 58.73 & 59.21 & \ul{59.37} &59.19 & 57.81 & \textbf{63.03} \\
\midrule
\multirow{7}{*}{\makecell{\textbf{MIMIC-}\\\textbf{CXR}}} & \textbf{BLEU} & \cellcolor{gray!15}3.28 & 4.07 & 3.39 & 3.29 & 3.53 & 3.99 & 3.14 & 3.34 & 3.62 & 3.37 & \ul{4.32} & \textbf{4.56} \\
& \textbf{ROUGE-L} & \cellcolor{gray!15}16.54 & \ul{18.75} & 17.25 & 16.14 & 16.58 & 18.62 & 15.52 & 16.79 & 16.67 & 15.75 & 18.64 & \textbf{19.03} \\
& \textbf{METEOR} & \cellcolor{gray!15}17.90 & 18.81 & 17.36 & 17.98 & 18.68 & 18.68 & 14.70 & 18.29 & 18.22 & 17.03 & \ul{19.78} & \textbf{20.23} \\
& \textbf{BERTScore} & \cellcolor{gray!15}85.57 & \ul{86.14} & 85.78 & 85.42 & 85.57 & \ul{86.14} & 84.75 & 85.59 & 85.58 &85.27 & 86.10 & \textbf{86.17} \\
& \textbf{CheXbert} & \cellcolor{gray!15}22.14 & 24.24 & 23.05 & 21.76 & 23.38 & \ul{25.14} & 20.07 & 22.74 & 22.95 & 22.76 & \textbf{25.78} & 24.93 \\
& \textbf{RadGraph} & \cellcolor{gray!15}9.43 & 10.73 & 9.78 & 9.20 & 9.93 & 10.75 & 7.74 & 9.52 & 9.93 & 9.40 & \ul{11.17} & \textbf{11.55} \\
& \textbf{RaTEScore} & \cellcolor{gray!15}40.00 & \ul{41.06} & 38.59 & 39.26 & 40.95 & 40.73 & 35.72 & 40.08 & 39.87 & 39.62 & 41.03 & \textbf{42.73} \\
\bottomrule
\end{tabular}}
\vspace{-0.1in}
\end{table*}

\noindent\ul{\textbf{Metrics.}}
\label{exp:metrics} \textbf{(1) Metrics for Performance Evaluation.}
For \textit{medical VQA}, we report Accuracy for close-ended questions and Recall for open-ended questions, following LLaVA-Med~\cite{li2024llava}.  
For \textit{medical report generation}, we use standard metrics for generation tasks, including BLEU~\cite{papineni2002bleu}, ROUGE-L~\cite{lin2004rouge}, METEOR~\cite{banerjee2005meteor}, and BERTScore~\cite{zhang2019bertscore}. Additionally, we report domain-specific metrics designed for medical report generation: CheXbert~\cite{smit2020combining}, RadGraph~\cite{jain2021radgraph} and RaTEScore~\cite{zhao2024ratescore}. More details of these metrics are provided in Appendix~\ref{appd:metric_report}.  \textbf{(2) Metrics for Attention Maps Evaluation.} For datasets with ground truth RoIs (e.g., SLAKE), we design two metrics to evaluate the attention distribution on images: (1) Coverage Score (Coverage), which measures spatial alignment, and (2) Intensity Alignment (Intensity), which assesses the degree of focus. Details of these two metrics are in Appendix~\ref{appd:metric_attention}.

\subsection{Medical VQA Results}

The performance of \ours on diverse medical VQA benchmarks is presented in Table~\ref{tab:main_exp_vqa}, indicating \ours maintains its superiority across all datasets and its effectiveness across diverse medical images and VQA tasks.
In addition, we also analyze visual attention distribution on the test set of SLAKE with annotated RoIs (Table~\ref{tab:interpret}). Since the baselines in Table~\ref{tab:main_exp_vqa} improve the model only on the decoding side without modifying attentions, their attention distributions remain similar to the base model, LLaVA-Med and we compare only against LLaVA-Med and LLaVA-Med + LoRA. We observe that \ours achieves the highest scores in both coverage (0.275) and intensity (0.147), outperforming baselines. These results highlight \ours's ability to focus more effectively on RoIs in medical images, explaining its better VQA performance and improved interpretability.

% We compare \ours with a range of inference-time baselines applied to the original \textbf{LLaVA-Med} and they show limited improvements. \ours achieves significant gains in accuracy by integrating visual attention tuning during fine-tuning. To ensure a fair comparison, we also evaluate all baselines on the fine-tuned \textbf{LLaVA-Med + LoRA}. While fine-tuning narrows the performance gap, \ours maintains its superiority across all datasets, demonstrating its effectiveness across diverse medical images and VQA tasks.
% However, in open-ended VQA tasks for VQA-RAD, some baselines such as DoLa perform slightly better, indicating room for improvement in handling complex free-text responses. 

\subsection{Medical Report Generation Results}
\label{sec:report_results}
Table~\ref{tab:main_exp_report_02} presents the evaluation of \ours on medical report generation tasks using traditional metrics (e.g., BLEU, ROUGE-L) and domain-specific metrics such as RaTEScore.  The baselines are applied on fine-tuned LLaVA-Med using LoRA, as the original LLaVA-Med performs significantly worse on this task, as shown in the full results in Appendix~\ref{appd:full_report_generation}.
Similar to the results in Table~\ref{tab:main_exp_vqa}, \ours outperforms all baselines across both datasets and almost all metrics. 
% On IU-Xray, it achieves substantial improvements, with a BLEU score of 11.05 and a RaTEScore of 63.03, surpassing baselines in both language quality and clinical accuracy. On MIMIC-CXR, \ours also shows the highest scores in most metrics. These results highlight the strength of our attention alignment and \moe design in report generation, which enhances both textual coherence and clinical potential in Med-LVLMs.

\subsection{Module Effectiveness Analysis}
\label{sec:ablation}

% In the design of \ours, we introduce a set of key components, including weak labels generation, selection of ``visually-critical'' heads and the \moe. The following experiments validate the effectiveness of these designs.

\subsubsection{Weak Labels Generation}
\label{sec:exp_quality_labels}

% we investigate the potential performance improvements with high-quality labels
\noindent\ul{\textbf{(1) Quality of Weak Labels.}}
This experiment evaluates the upper bound of \ours 's performance by upgrading weak labels to high-quality ground truth labels (\textbf{GT}). We use a subset of SLAKE with RoIs annotations, including 992 training samples and 206 test samples and take ControlMLLM~\cite{wu2024controlmllm} as the  baseline.

% \begin{table}[h!]
% \centering
% \caption{Performance comparison with high-quality segmentation labels on SLAKE. Results are reported for open- and close-ended tasks. \textbf{Base} indicates the base model, \textbf{+Control} adds ControlMLLM, \textbf{+Weak} uses weak labels, and \textbf{+GT} uses ground truth labels.}
% \label{tab:high_quality_Segs}
% \vspace{-0.1in}
% \resizebox{\columnwidth}{!}{
% \begin{tabular}{l|cc|cc|cc}
% \toprule
% \textbf{VQA Type} & \multicolumn{2}{c|}{\textbf{LLaVA-Med}} & \multicolumn{2}{c|}{\makecell{ \textbf{LLaVA-Med + }\\ \textbf{LoRA}}} & \multicolumn{2}{c}{\makecell{ \textbf{LLaVA-Med + }\\ \textbf{\ours}}}  \\ 
% \cline{2-7}
% & \textbf{Base} & \textbf{+Control} & \textbf{Base} & \textbf{+Control} & \textbf{+Weak} & \textbf{+GT} \\
% \midrule
% Open-ended   & 31.39 & 33.04 & 72.12  & 70.61& 71.37 & 72.84 \\
% Close-ended  & 63.46 & 67.31 & 84.62 & 84.62& 86.54 & 88.46 \\
% \bottomrule
% \end{tabular}}
% \vspace{-0.1in}
% \end{table}

\begin{figure}[t]
  \centering
  \includegraphics[width=0.95\linewidth]{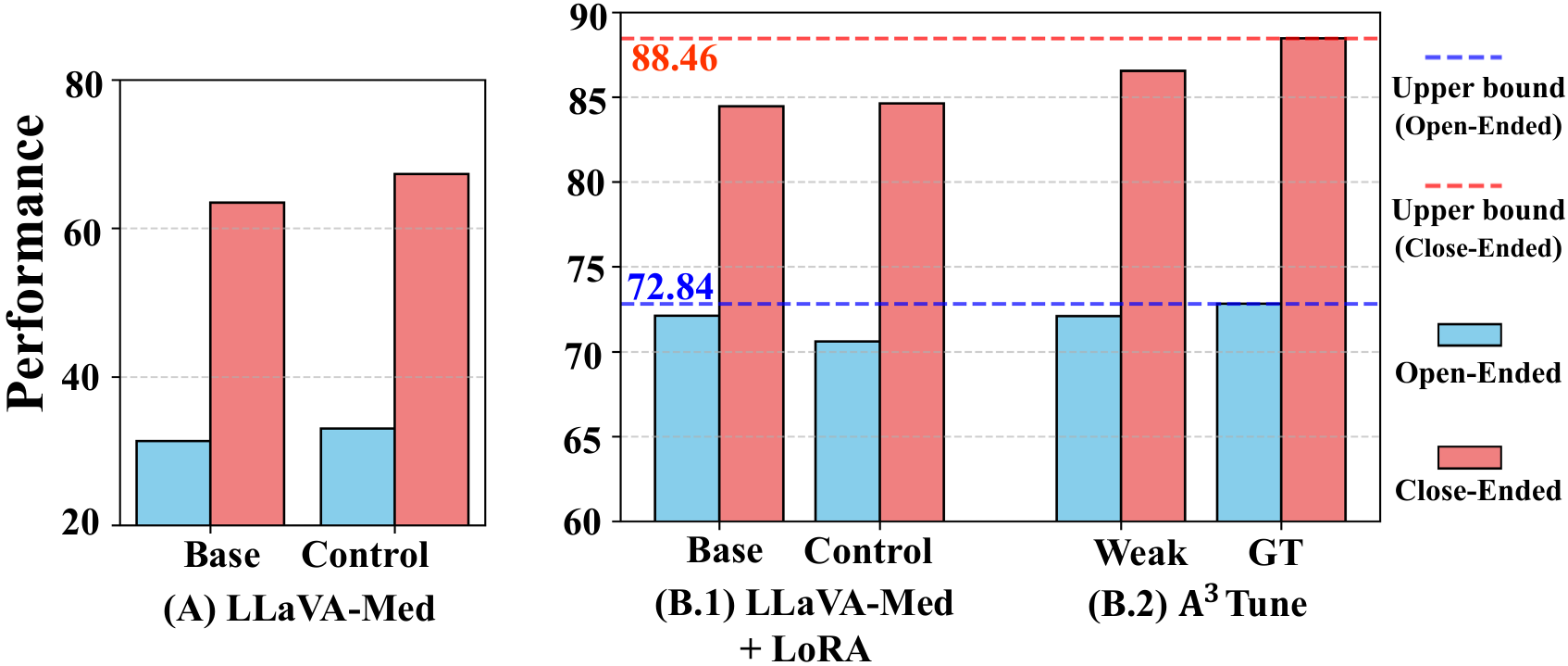}
  \vspace{-0.1in}
  \caption{Effectiveness analysis of RoIs labels. \textbf{Base} is the base model, \textbf{Control} means adding ControlMLLM to align attention maps with ground truth labels, \textbf{Weak} uses weak labels, and \textbf{GT} uses ground truth labels.}
  \label{fig:upper_bound_analysis}
  \vspace{-0.15in}
\end{figure}

Figure~\ref{fig:upper_bound_analysis} (B.2) shows that replacing weak labels (\textbf{Weak}) with high-quality labels (\textbf{GT} in \ours further improves performance, validating the reasonableness of using weak labels. Additionally, while ControlMLLM (\textbf{Control}) improves performance in LLaVA-Med (Figure~\ref{fig:upper_bound_analysis} (A)), it negatively impacts fine-tuned LLaVA-Med+LoRA in Figure~\ref{fig:upper_bound_analysis} (B.1), whereas \ours achieve improved results. These results highlight both the effectiveness of \ours and its potential for further improvement when ground truth RoI labels are available.

\begin{figure}[t]
  \centering
  \includegraphics[width=0.95\linewidth]{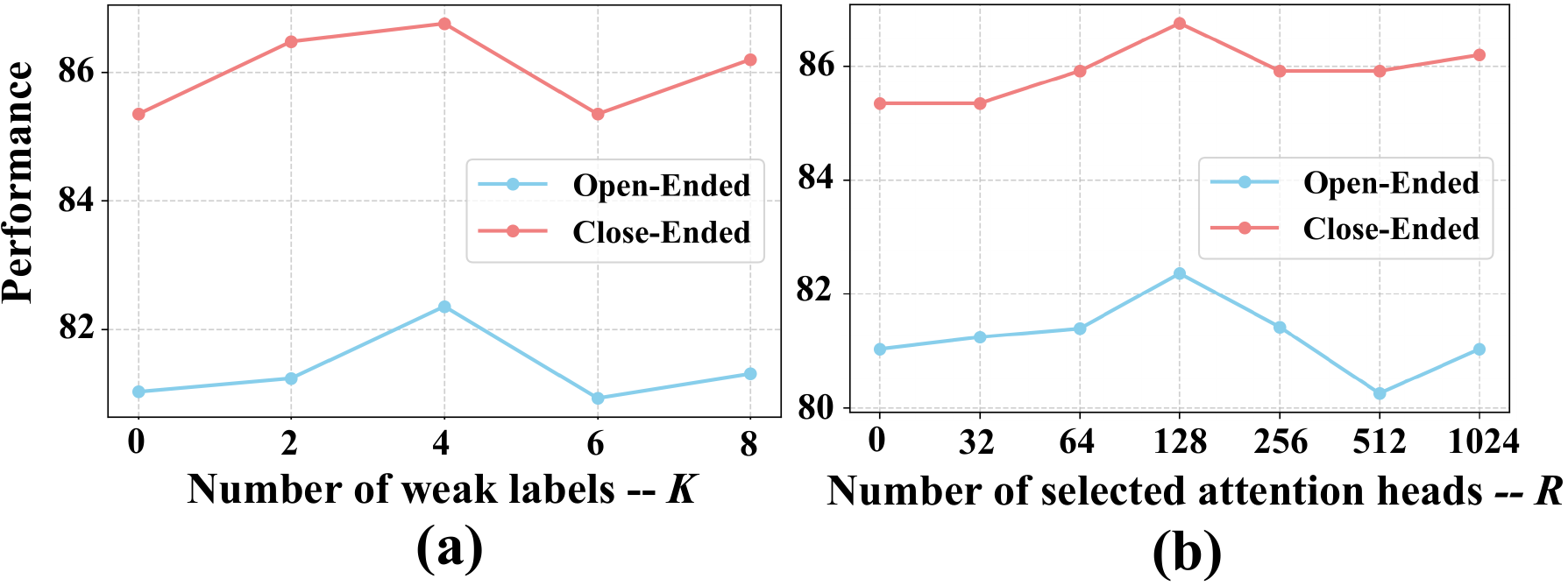}
  \vspace{-0.1in}
  \caption{Analysis of (a) the number of selected attention heads $R$ in \ours and (b) the number of weak labels $K$, evaluated on the SLAKE dataset.}
  \label{fig:hyper_param_analysis}
  \vspace{-0.15in}
\end{figure}

\begin{figure*}[t]
  \centering
  \includegraphics[width=0.95\linewidth]{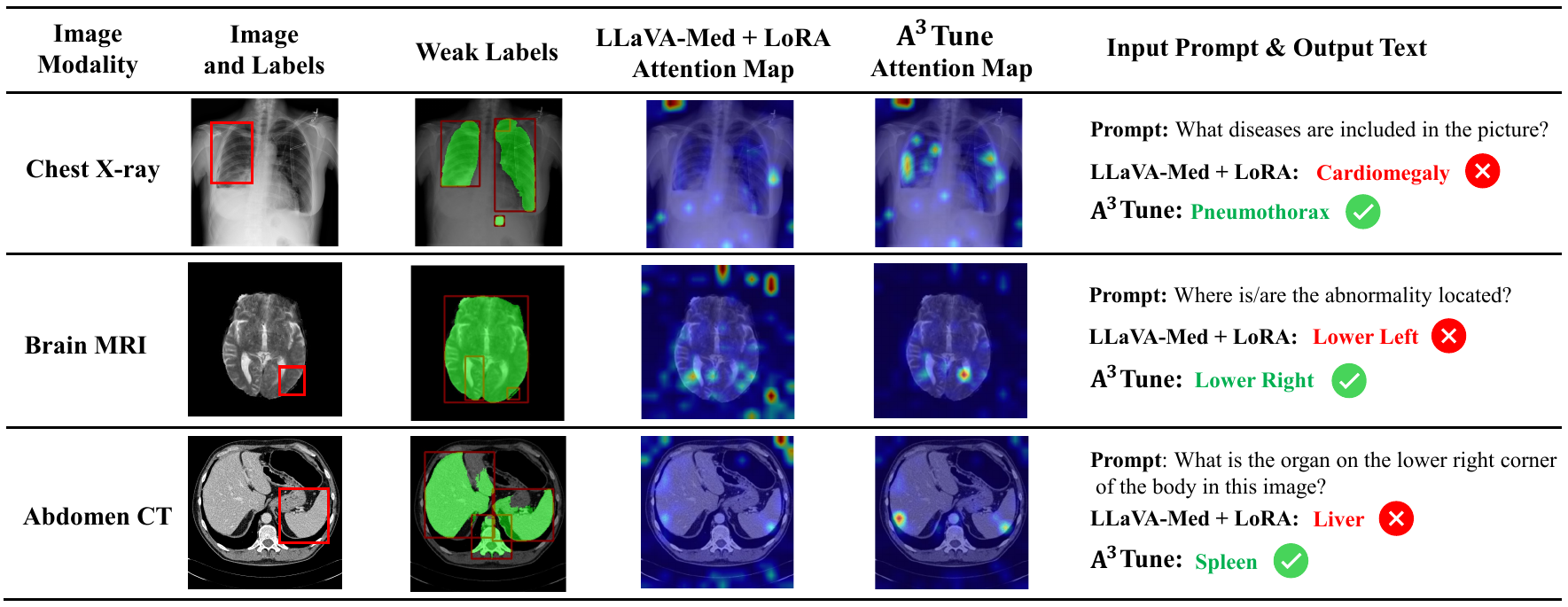}
  \vspace{-0.1in}
  \caption{Case study for fine-grained effectiveness 
 analysis. The \textcolor{red}{red box} in the first column (not provided as input to the model) highlights the RoI that LLaVA-Med should focus on. The second column shows weak segmentation labels, with red bounding boxes and green inner segments generated using the method described in Section~\ref{sec:method_weak_label}. 
 % In the output text, incorrect answers are highlighted in red, and correct answers are highlighted in green.
 }
  \label{fig:case_study}
  \vspace{-0.15in}
\end{figure*}

\begin{figure}[t]
  \centering
  \includegraphics[width=0.95\linewidth]{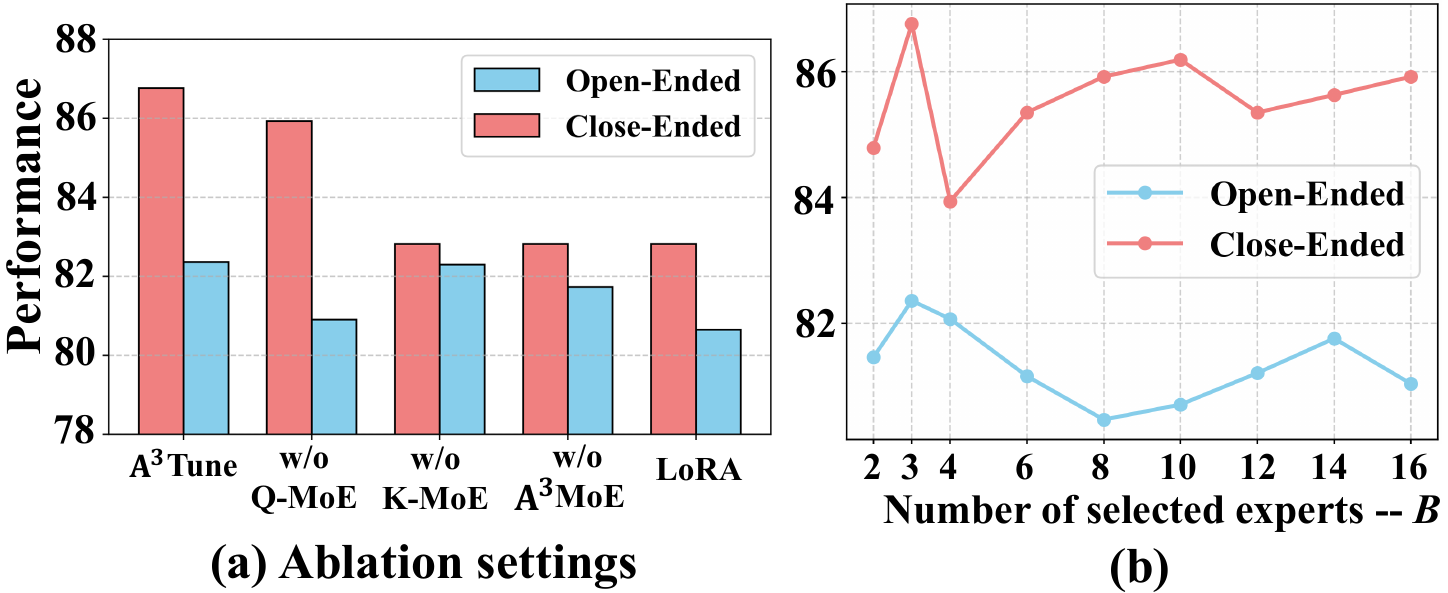}
  \vspace{-0.1in}
  \caption{Analysis of (a) the contribution of different modules in \moe, and (b) the number of experts in K-MoE within \moe. In (a), each module's impact is evaluated by removing it from \ours, denoted "w/o".}
  \label{fig:moe_effective_analysis}
  \vspace{-0.15in}
\end{figure}

\noindent\ul{\textbf{(2) Selection of Weak Labels.}}
To obtain prompt-aware weak labels, we select $K$ segments that are most similar to the text prompt. In this experiment, we analyze the impact of different values of $K$ on model performance using the SLAKE dataset. As shown in Figure~\ref{fig:hyper_param_analysis} (a)
increasing $K$ initially improves performance by adding more relevant labels. However, performance peaks at $K=4$, after which additional segments introduce noise, leading to a performance decline.

\subsubsection{Selection of Attention Heads}
\label{subsec:selection_k}
In \ours, we select $R$ most ``visually-critical'' attention heads to balance model stability and effectiveness of \ours. 
% This experiment explores how different values of $R$ affect the model’s performance.  Notably, 
% $R=0$ represents the removal of attention alignment tuning, while $R=1024$ equals tuning all attention heads in LLaVA-Med. 
As shown in Figure~\ref{fig:hyper_param_analysis} (b), disabling attention tuning ($R=0$) results in noticeably lower performance. Performance peaks at $R=128$, which strikes a balance between the strength of attention tuning and stability. Thus, we set $R=128$ in our experiments.
% Beyond this point, performance declines, likely due to excessive intervention across too many attention heads.

\subsubsection{\moe Design}

\begin{table*}[t]
\centering
\caption{Performance on SLAKE VQA benchmark using HuatuoGPT-Vision-7B (denoted as \textbf{HuatuoGPT-V}). We report both open-ended and close-ended performance.}
\label{tab:huatuo_exp_vqa}
\vspace{-0.1in}
\resizebox{2\columnwidth}{!}{
\begin{tabular}{p{2.3cm}|p{2cm}|ccccccccccc|c}
\toprule
\multirow{2}{*}{\textbf{Model}} & \multirow{2}{*}{\textbf{Metric}} & \multicolumn{12}{c}{\textbf{Method}} \\
\cline{3-14}
& & \cellcolor{gray!15}\makecell{Base} 
& Greedy 
& Beam 
& Nucleus 
& VCD 
& DoLa 
& OPERA 
& AVISC 
& M3ID 
& DAMRO 
& PAI 
& \textbf{A$^3$Tune} \\
\midrule
% \multirow{2}{*}{\textbf{HuatuoGPT-V}} 
% & Open-ended      & \cellcolor{gray!15}--    & 51.97 & 50.98 & 50.62 & 51.82 & 49.61 & 52.35 & \ul{54.89} & 53.88 & 51.82 & 54.50 & -- \\
% & Close-ended     & \cellcolor{gray!15}--    & 71.27 & 72.39 & 68.73 & 70.14 & 66.76 & 72.68 & \ul{77.18} & 67.04 & 70.14 & 71.55 & -- \\
% \midrule
\multirow{2}{*}{\makecell{\textbf{HuatuoGPT-V}\\\textbf{+ LoRA}}} 
& Open-ended      & \cellcolor{gray!15}85.46 & 84.89 & \ul{85.62} & 85.57 & 85.57 & 85.19 & 85.07 & 84.42 & 85.41 & 85.57 & 83.85 & \textbf{86.77} \\
& Close-ended     & \cellcolor{gray!15}\ul{91.27} & 89.86 & 89.86 & 90.99 & 90.99 & 90.42 & 89.86 & 90.70 & \ul{91.27} & 90.99 & 90.99 & \textbf{91.55} \\
\bottomrule
\end{tabular}
}
\vspace{-0.1in}
\end{table*}

\begin{table*}[t]
\centering
\caption{Performance on report generation benchmarks using HuatuoGPT-Vision-7B fine-tuned with LoRA.}
\label{tab:huatuo_exp_report}
\vspace{-0.1in}
\resizebox{2\columnwidth}{!}{
\begin{tabular}{p{1.3cm}|p{1.8cm}|ccccccccccc|c}
\toprule
\multirow{2}{*}{\textbf{Dataset}} & \multirow{2}{*}{\textbf{Metric}} & \multicolumn{12}{c}{\textbf{Method}}\\
\cline{3-14}
& & \cellcolor{gray!15}\makecell{HuatuoGPT-V\\ + LoRA} 
& Greedy 
& Beam
& Nucleus
& VCD
& DoLa
& OPERA
& AVISC
& M3ID
& DAMRO
& PAI
& \ours \\
\midrule
\multirow{7}{*}{\textbf{IU-Xray}} 
& \textbf{BLEU}      & \cellcolor{gray!15}8.65  & 9.34  & \ul{10.21} & 8.19  & 9.10  & 9.03  & 10.01 & 7.47  & 8.35  & 9.10  & 6.92  & \textbf{10.52} \\
& \textbf{ROUGE-L}   & \cellcolor{gray!15}27.34 & 28.17 & \ul{28.64} & 26.28 & 27.80 & 27.50 & 28.57 & 24.78 & 27.05 & 27.80 & 24.36 & \textbf{28.85} \\
& \textbf{METEOR}    & \cellcolor{gray!15}31.48 & 31.76 & \ul{34.23} & 30.72 & 32.13 & 31.24 & 34.10 & 30.88 & 32.17 & 32.13 & 30.84 & \textbf{36.30} \\
& \textbf{BERTScore} & \cellcolor{gray!15}88.35 & 88.53 & \ul{88.60} & 88.19 & 88.36 & 88.39 & 88.51 & 87.77 & 88.19 & 88.36 & 87.44 & \textbf{88.67} \\
& \textbf{CheXbert}  & \cellcolor{gray!15}54.21 & 55.16 & \ul{55.84} & 53.86 & 54.34 & 53.89 & 55.01 & 52.11 & 53.78 & 54.34 & 50.09 & \textbf{56.27} \\
& \textbf{RadGraph}  & \cellcolor{gray!15}21.35 & 21.86 & 22.47 & 20.17 & 21.65 & 21.04 & \ul{22.59} & 19.71 & 21.19 & 21.65 & 18.26 & \textbf{23.51} \\
& \textbf{RaTEScore} & \cellcolor{gray!15}58.21 & 58.66 & \ul{59.78} & 58.29 & 58.29 & 57.84 & 59.33 & 56.49 & 57.86 & 58.29 & 55.16 & \textbf{60.51} \\
\bottomrule
\end{tabular}}
\vspace{-0.1in}
\end{table*}

\noindent\ul{\textbf{(1) Ablation Study of \moe.}}
Figure~\ref{fig:moe_effective_analysis} (a) shows the impact of removing key components of \moe from \ours: Q-MoE, K-MoE and \moe as a whole. The removal of each module leads to a noticeable performance drop, particularly for K-MoE. This highlights the importance of K-MoE, which is designed for fine-grained attention tuning at the visual token level and is critical for datasets with diverse image modalities like SLAKE. Additionally, the removal of the entire \moe leads to a further decline in performance. However, even with these components removed, the performance of \ours remains above the baseline LoRA tuning on LLaVA-Med, demonstrating the effectiveness of the overall framework.

\noindent\ul{\textbf{(2) The Number of Selected Experts in K-MoE.}} As shown in Figure~\ref{fig:moe_effective_analysis}(b), performance peaks at $B=3$. Beyond this point, additional experts lead to diminishing and unstable returns, likely due to saturation and increased interference among experts handling different visual tokens.

\subsection{Case Study}
\label{sec:case_study}

Figure~\ref{fig:case_study} presents a case study for three image modalities. The attention map visualizations demonstrate that our method effectively redirects the model's focus to relevant regions, mitigating hallucination issues compared to baselines. In addition, we provide fine-grained effectiveness evaluation across Chest X-ray, Abdomen CT, and Brain MRI images from SLAKE in Appendix~\ref{appd:finegrained_effectiveness} Table~\ref{tab:med_modality}.

\subsection{Generalization to Other Medical LVLMs}
We further evaluate our approach on medical VQA (Table~\ref{tab:huatuo_exp_vqa}) and report generation (Table~\ref{tab:huatuo_exp_report}) using a more recent and stronger Med-LVLM, HuatuoGPT-Vision-7B~\cite{chen2024huatuogptv}. As shown in the results, transferring \ours to this new backbone consistently achieves the best performance across all metrics on IU-Xray for report generation and SLAKE for VQA. These results demonstrate the flexibility and strong generalization ability of \ours, along with its capacity to further enhance performance when built upon more capable backbone models.

\section{Conclusion}
In this work, we present \ours, a novel fine-tuning framework designed to enhance the visual grounding capabilities of Med-LVLMs. By leveraging prompt-aware weak labels and integrating a \moe design, \ours dynamically aligns attention distributions to RoIs across diverse medical tasks and datasets, without requiring inference-time adjustments. Extensive experiments 
% demonstrate that \ours improves both performance and interpretability in medical VQA and report generation over baselines. 
% These results 
highlights \ours as a promising direction for enhancing  Med-LVLMs in downstream applications.

% \newpage

\section*{Limitations}

While the use of weak labels in \ours demonstrates its effectiveness, it also introduces noise that can limit performance. In some cases, the model can only focus on generally correct regions, lacking accuracy but providing directions for future research. As shown in Section~\ref{sec:exp_quality_labels}, using high-quality labels leads to further performance improvements. Additionally, our framework is currently restricted to fine-tuning for downstream tasks, limiting its broader applicability. Furthermore, the metrics used to evaluate visual attention distribution are constrained to patch-level granularity due to the inherent design of Med-LVLMs, rather than achieving pixel-level precision.

% \section*{Ethics Statement}

\section*{Acknowledgments}
The authors thank the anonymous referees for their valuable comments and helpful suggestions. This work is partially supported by the National Science Foundation under Grant No. 2238275 and the National Institutes of Health under Grant No. R01AG077016.

\bibliography{custom}
\bibliographystyle{acl_natbib}

\appendix
% \clearpage

% \section{Experiment Settings and Hyperparameters}
% % \subsection{LVLM Settings}

\section{Dataset Processing and Statistics}
\label{appd:dataset}
\begin{table}[h]
\centering
\resizebox{0.9\columnwidth}{!}{
\begin{tabular}{c|c|c|c}
\hline
\textbf{Task}          & \textbf{Dataset} & \textbf{Training} & \textbf{Test} \\ \hline
\multirow{5}{*}{\makecell{Medical \\ VQA} } 
& SLAKE           & 4,919                     & 1,061        \\ \cline{2-4} 
& VQA-RAD         & 1,797                     & 451  \\ \cline{2-4} 
& PathVQA         & 19,755                     & 6,761    \\ 
\cline{2-4} 
& IU-Xray         & 1,789                     & 784     \\ 
\cline{2-4} 
& OmniMedVQA         & 6,155                     & 2,642  \\ 
\hline
\multirow{2}{*}{\makecell{Report \\ Generation} } 
& IU-Xray         & 2,069                     & 590        \\ \cline{2-4} 
& MIMIC-CXR       & 1,902                     & 441       \\ \hline
\end{tabular}}
\caption{Dataset statistics for the Medical VQA and Report Generation tasks.}
\label{tab:dataset_stats}
\end{table}
For IU-Xray and OmniVQA in medical VQA task, we utilize the preprocessed datasets provided by the CARES benchmark~\cite{xia2024cares}, splitting each dataset into training and test sets with a 7:3 ratio. For the MIMIC-CXR dataset used in the report generation task, we randomly sample 2,000 image-report pairs from the preprocessed MIMIC-CXR-JPG dataset~\cite{johnson2019mimic} for the training set and 500 pairs for the test set. We extract the ``Findings'' and ``Impression'' sections from each report in sampled MIMIC-CXR reports, filtering out those with an extremely low word count. The statistics of those datasets are shown in Table~\ref{tab:dataset_stats}.

\section{Implementation Details of Baselines}
\label{appd:implementation_baselines}
Generally, we follow the recommended settings for all baselines while making necessary adjustments to adapt them to Med-LVLMs. The detailed settings are listed as follows:
\begin{itemize}  
    \item Beam Search~\cite{NIPS2014_beam}: The number of beams is set to 5.  
    \item Nucleus Sampling~\cite{Holtzman2020The}: The top-\( p \) value for sampling is \( 0.9 \).  
    \item VCD~\cite{leng2024mitigating}: The contrastive decoding parameters are set to \( \alpha = 1 \) and \( \beta = 0.1 \). Diffusion noise is added to images using 500 steps.  
    \item DoLa~\cite{chuang2023dola}: The mature layer is set to 32, while the early candidate mature layers are \([0,2,4,6,8,10,12,14]\).  
    \item OPERA~\cite{huang2024opera}: The number of beams is set to 5, with a scale factor of 50, threshold of 15, and \(\text{num-attn-candidates} = 5\). The penalty weight is set to 1. Notably, for LLaVA-Med-1.5 in the report generation task, the scale factor is set to 25 and the threshold is adjusted to 25, as the default values result in nonsensical decoded content.  
    \item AVISC~\cite{woo2024don}: We select the top-10 outlier image tokens to construct the negative decoding object. The contrastive decoding parameters are set to \( \alpha = 1 \) and \( \beta = 0.1 \).
    \item M3ID~\cite{favero2024multi}: The contrastive decoding parameters are set as follows: $\lambda = 0.02$ and $\gamma_t = \exp(-\lambda \cdot t)$, where $t$ denotes the current decoding step.
    \item DAMRO~\cite{gong2024damro}: We select the top-10 tokens with the highest attention to the [CLS] token in the visual encoder as outlier tokens. The contrastive decoding parameters are set to \( \alpha = 0.5 \) and \( \beta = 0.1 \).
    \item PAI~\cite{liu2025paying}: In the inference intervention, the start layer and end layer are set to 2 and 32, respectively, \(\gamma = 1.1\) and \(\alpha = 0.2\).  
    \item ControlMLLM~\cite{wu2024controlmllm}: In inference-time tuning, it is configured with $\beta=0.5\text{, } \alpha=400$, and a learning rate of $4$. For LLaVA-Med-1.5, the same parameters are applied but with a reduced learning rate of $1$.
\end{itemize}  

\section{Implementation Details}
\label{appd:implementation}
\begin{table}[t]
    \centering
    \caption{Fine-tuning epochs and $\lambda$ values for different datasets.}
    \label{tab:hyperparams}
    \resizebox{0.6\columnwidth}{!}{\begin{tabular}{lcc}
        \toprule
        \textbf{Dataset} & \textbf{Epochs} & \boldmath$\lambda$ \\
        \midrule
        \multicolumn{3}{c}{\textit{Medical VQA}} \\
        \midrule
        SLAKE         & 6  & 0.1  \\
        VQA-RAD       & 9  & 0.06 \\
        PathVQA       & 3  & 0.02 \\
        IU-Xray       & 6  & 0.12 \\
        OmniMedVQA    & 3  & 0.03 \\
        \midrule
        \multicolumn{3}{c}{\textit{Medical Report Generation}} \\
        \midrule
        IU-Xray       & 12 & 0.08 \\
        MIMIC-CXR     & 12 & 0.05 \\
        \bottomrule
    \end{tabular}}
    \label{tab:hyper_lambda_epoch}
\end{table}

\subsection{Hyperparameter setting}
All fine-tuning tasks are performed using the same seed for LoRA initialization. The LoRA rank is set to 64 and the rank of each expert in \moe is set to 16, with a learning rate of 2e-4. For \moe, the default number of experts in K-MoE $O^k$ is 8, while the default number of experts in Q-MoE $O^q$ is 4. For example, we use these settings in all report generation tasks for Chest X-rays. However, for large datasets with diverse image modalities such as SLAKE, PathVQA and OmniMedVQA, the number of experts is increased to 16 and 8, respectively. Some other key hyperparameters are: $K=4$, $R=128$, $B=3$ when $O^k=16$ and $B=2$ when $O^k=8$.

\begin{figure}[h!]
  \centering
  \includegraphics[width=0.9\linewidth]{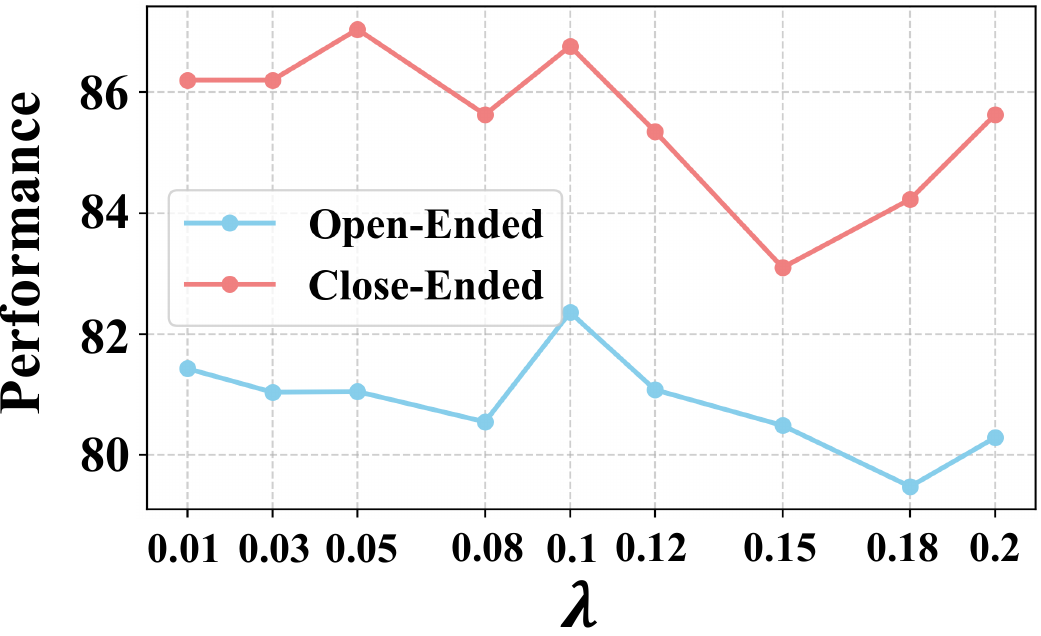}
  \vspace{-0.1in}
  \caption{Analysis of the value of $\lambda$ in \ours on the SLAKE dataset.}
  \label{fig:hyper_lambda}
  \vspace{-0.15in}
\end{figure}

\begin{table*}[t]
\centering
\caption{Full results of report generation on IU-Xray, based on LLaVA-Med and the LoRA fine-tuned LLaVA-Med.}
\label{tab:report_app01}
\vspace{-0.1in}
\resizebox{1.85\columnwidth}{!}{
\begin{tabular}{c|c|ccccccc}
\toprule
\multirow{2}{*}{\textbf{Model}} & \multirow{2}{*}{\textbf{Method}} & \multicolumn{7}{c}{\textbf{IU-Xray}}\\
\cline{3-9}
 & &\textbf{BLEU} & \textbf{\text{ROUGE-L}} & \textbf{METEOR} & \textbf{BERTScore} & \textbf{CheXbert} & \textbf{RadGraph} & \textbf{RaTEScore}\\
\midrule
% \multicolumn{8}{c}{\textit{Method based on LLaVA-Med Without Fine-tuning}} \\
% \midrule
\multirow{11}{*}{\textbf{LLaVA-Med}} & 
 \cellcolor{gray!15}LLaVA-Med & \cellcolor{gray!15}1.30 & \cellcolor{gray!15}11.55 & \cellcolor{gray!15}16.09 & \cellcolor{gray!15}84.12 & \cellcolor{gray!15}35.21 & \cellcolor{gray!15}6.46 & \cellcolor{gray!15}42.32  \\
& Greedy  & 1.21 & 13.97 & 19.12 & 84.61 & 35.82 & 6.74 &42.32   \\
& Beam & 1.19 & 13.42 & 19.49 & 84.24 & 39.41 & 8.69 & 44.56 \\
& Nucleus & 1.33 & 11.15 & 16.28 & 83.89 & 32.39 & 6.39 & 41.39 \\
& VCD & 1.19 & 10.26 & 15.46 & 83.49 & 33.25 & 5.67 & 41.20 \\
& DoLa & 1.20 & 13.93 & 19.14 & 84.60 & 35.61 & 6.74 & 42.31  \\
& OPERA & 0.89 & 8.46 & 13.90 & 82.52 & 33.19 & 3.48 & 39.19 \\
& AVISC & 1.16 & 11.12 & 17.34 & 83.86 & 30.99 & 5.08 & 43.30  \\
& M3ID & 1.40 & 12.01 & 16.48 & 84.07 & 34.32 & 6.71 & 42.02 \\
& DAMRO & 1.23 & 10.69 & 16.68 & 83.72 & 30.21 & 5.63 & 43.62 \\
& PAI & 1.14 & 12.90 & 18.46 & 84.33 & 38.64 & 6.96 & 42.92 \\
\midrule
% \multicolumn{8}{c}{\textit{Method based on LLaVA-Med With LoRA Fine-tuning}} \\
% \midrule
\multirow{12}{*}{\textbf{\makecell{LLaVA-Med \\ + LoRA}}} & 
\cellcolor{gray!15}LLaVA-Med + LoRA & \cellcolor{gray!15}7.70 & \cellcolor{gray!15}26.15 & \cellcolor{gray!15}29.50 &\cellcolor{gray!15}88.36 & \cellcolor{gray!15}53.81  & \cellcolor{gray!15}20.44 & \cellcolor{gray!15}58.37 \\
& Greedy  & 8.86 & 27.09 & 26.01 & 88.50 & 52.55 & 20.76 &58.24   \\
& Beam & \ul{9.34} & 27.56 & 26.44 & \ul{88.52} & 52.88 & 21.29 & 58.77 \\
& Nucleus & 7.80 & 26.72 & 30.33 & 88.28 & 52.73 & 20.85 & 57.84 \\
& VCD & 8.83 & 27.36 & \ul{31.77} & 88.30 & 51.86 & 22.02 & 58.93 \\
& DoLa & 8.93 & 26.94 & 25.74 & 88.42 & 52.27 & 20.63 & 58.10  \\
& OPERA & 8.27 & 27.14 & 29.66 & 88.41 & \ul{56.26} & 21.87 & 58.73 \\
& AVISC & 8.52 & 26.97 & 31.14 & 88.47 & 53.33 & \ul{22.27} & 59.21  \\
& M3ID & 8.63 & \ul{27.79} & 31.65 & \ul{88.52} & 54.45 & 22.22 & \ul{59.37} \\
& DAMRO & 7.29 & 25.61 & 30.03 & 88.14 & 51.50 & 20.75 & 59.19 \\
& PAI & 8.87 & 26.74 & 25.99 & 88.39 & 52.45 & 20.48 & 57.81  \\
% \midrule
& \ours(ours) & \textbf{11.05} & \textbf{30.00} & \textbf{34.26} & \textbf{89.05} & \textbf{57.19} & \textbf{24.24} & \textbf{63.03} \\
\bottomrule
\end{tabular}}
% \vspace{-0.1in}
\end{table*}

\begin{table*}[h]
\centering
\caption{Full results of report generation on MIMIC-CXR, based on LLaVA-Med and the LoRA fine-tuned LLaVA-Med.}
\label{tab:report_app02}
\vspace{-0.1in}
\resizebox{1.8\columnwidth}{!}{
\begin{tabular}{c|c|ccccccc}
\toprule
\multirow{2}{*}{\textbf{Model}} & \multirow{2}{*}{\textbf{Method}} & \multicolumn{7}{c}{\textbf{MIMIC-CXR}}\\
\cline{3-9}
 & & \textbf{BLEU} & \textbf{\text{ROUGE-L}} & \textbf{METEOR} & \textbf{BERTScore} & \textbf{CheXbert} & \textbf{RadGraph} & \textbf{RaTEScore}\\
\midrule
% \multicolumn{8}{c}{\textit{Method based on LLaVA-Med With LoRA Fine-tuning}} \\
% \midrule
\multirow{11}{*}{\textbf{LLaVA-Med}} & 
 \cellcolor{gray!15}LLaVA-Med & \cellcolor{gray!15}1.38 & \cellcolor{gray!15}12.28 & \cellcolor{gray!15}13.20 & \cellcolor{gray!15}84.24 & \cellcolor{gray!15}15.45 & \cellcolor{gray!15}3.14 & \cellcolor{gray!15}32.91 \\
& Greedy  & 0.42 & 9.23 & 6.77 & 83.94 & 14.23 & 1.42 &29.81   \\
& Beam & 1.20 & 13.06 & 13.00 & 83.25 & 10.90 & 3.16 & 34.30 \\
& Nucleus & 1.22 & 12.49 & 11.99 & 83.65 & 15.79 & 3.31 & 34.17 \\
& VCD & 1.12 & 12.21 & 11.53 & 83.53 & 14.25 & 2.83 & 34.06 \\
& DoLa & 0.39 & 9.13 & 6.54 & 83.96 & 14.26 & 1.38 & 29.59  \\
& OPERA & 1.06 & 12.90 & 11.93 & 83.11 & 12.90 & 2.80 & 36.12 \\
& AVISC & 1.20 & 11.90 & 13.08 & 83.20 & 13.09 & 2.62 & 35.22  \\
& M3ID & 1.10 & 12.13 & 11.47 & 83.69 & 15.08 & 2.77 & 33.68 \\
& DAMRO & 1.27 & 12.40 & 12.84 & 83.40 & 13.76 & 3.36 & 35.71 \\
& PAI & 0.48 & 9.74 & 7.22 & 83.81 & 14.39 & 1.45 & 30.57 \\
\midrule
% \multicolumn{8}{c}{\textit{Method based on LLaVA-Med With LoRA Fine-tuning}} \\
% \midrule
\multirow{12}{*}{\textbf{\makecell{LLaVA-Med \\ + LoRA}}} & 
 \cellcolor{gray!15}LLaVA-Med + LoRA & \cellcolor{gray!15}3.28 & \cellcolor{gray!15}16.54 & \cellcolor{gray!15}17.90 &\cellcolor{gray!15}85.57 & \cellcolor{gray!15}22.14  & \cellcolor{gray!15}9.43 & \cellcolor{gray!15}40.00 \\
& Greedy  & 4.07 & \ul{18.75} & 18.81 & \ul{86.14} & 24.24 & 10.73 & \ul{41.06}   \\
& Beam & 3.39 & 17.25 & 17.36 & 85.78 & 23.05 & 9.78 & 38.59 \\
& Nucleus & 3.29 & 16.14 & 17.98 & 85.42 & 21.76 & 9.20 & 39.26 \\
& VCD & 3.53 & 16.58 & 18.68 & 85.57 & 23.38 & 9.93 & 40.95 \\
& DoLa & 3.99 & 18.62 & 18.68 & \ul{86.14} & \ul{25.14} & 10.75 & 40.73  \\
& OPERA & 3.14 & 15.52 & 14.70 & 84.75 & 20.07 & 7.74 & 35.72 \\
& AVISC & 3.34 & 16.79 & 18.29 & 85.59 & 22.74 & 9.52 & 40.08 \\
& M3ID & 3.62 & 16.67 & 18.22 & 85.58 & 22.95 & 9.93 & 39.87 \\
& DAMRO & 3.37 & 15.75 & 17.03 & 85.27 & 22.76 & 9.40 & 39.62 \\
& PAI & \ul{4.32} & 18.64 & \ul{19.78} & 86.10 & \textbf{25.78} & \ul{11.17} & 41.03  \\
% \midrule
& \ours (ours) & \textbf{4.56} & \textbf{19.03} & \textbf{20.23} & \textbf{86.17} & 24.93 & \textbf{11.55} & \textbf{42.73} \\
\bottomrule
\end{tabular}
}
% \vspace{-0.1in}
\end{table*}

\subsection{$\lambda$ Selection}
In our final loss in Eq.~\eqref{eq:lambda}, we use a key hyperparameter $\lambda$ to balance the two loss terms. Here, we conduct an analysis to select the optimal value of $\lambda$.
Figure~\ref{fig:hyper_lambda}, illustrate the impact of $\lambda$, which controls the strength of attention alignment tuning. We can observe that on SLAKE, performance peaks at $\lambda=0.1$. Beyond this point, the performance declines due to over-alignment and extreme attention distributions. Notably, $\lambda$ varies with the scale of datasets and the training epoches, and the values of $\lambda$ and fine-tuning epoches for all the datasets are as shown in Table~\ref{tab:hyper_lambda_epoch}. 

All the experiments are conducted using four A6000 GPUs.

\section{Metrics}
\label{appd:metrics}
\subsection{Metrics for Report Generation}
\label{appd:metric_report}
We evaluate model performance using commonly used metrics for generation tasks. These include BERTScore~\cite{zhang2019bertscore}, which measures the similarity between the embeddings of predicted and reference texts, and METEOR~\cite{banerjee2005meteor}, which evaluates alignment between generated answers and reference texts, accounting for synonyms and stemming. Additionally, we employ ROUGE-L~\citep{lin2004rouge}, which measures n-gram overlap and the longest common subsequence, and BLEU~\cite{papineni2002bleu}, which calculates n-gram precision in the predicted text relative to the reference, focusing on exact matches. In addition, we include the following domain-specific metrics designed for medical report generation: 
\begin{itemize}
    \item CheXbert~\cite{smit2020combining} is an automatic labeler that extracts pathology indicators from radiology reports. We follow~\cite{yu2023evaluating} to calculate the CheXbert vector similarity that measures the cosine similarity between pathology indicator vectors derived from ground truth and model-generated reports. 
    \item RadGraph~\cite{jain2021radgraph} is a tool that extracts entity and relation from radiology reports. We use RadGraph to specifically indicate RadGraph F1, which measures the overlap of clinical entities and their relations extracted from ground truth and model-generated reports.
    \item RaTEScore~\cite{zhao2024ratescore} is a recently proposed metric that prioritizes crucial medical entities, including diagnostic outcomes and anatomical details. This metric is robust to complex medical synonyms and sensitive to negation expressions, aligning more closely with human judgment compared to existing metrics.
\end{itemize}

\subsection{Metrics for Attention Distribution}
\label{appd:metric_attention}

\textbf{(a) Coverage Score.} 
The Coverage Score measures the proportion of the ground truth region that is covered by the attention map. Let $\mathbf{B}$ denote the binary mask of the ground truth segment (where $\mathbf{B}(i, j) = 1$ for pixels belonging to the ground truth and $\mathbf{B}(i, j) = 0$ otherwise) and $\mathbf{M}$ denote the attention map output by the model. The score is defined as:
$$
\text{Coverage} = \frac{\sum_{i, j} \mathbf{B}(i, j) \cdot \mathbf{M}_{\tau}(i, j)}{\sum_{i, j} \mathbf{B}(i, j)},
$$
where $\mathbf{M}_{\tau}$ is the thresholded attention map, i.e., $\mathbf{M}_{\tau}(i, j) = 1$ if $\mathbf{M}(i, j) \geq {\tau}$, and $\mathbf{M}_{\tau}(i, j) = 0$ otherwise. This metric quantifies how well the attention aligns spatially with the ground truth. In our experiments, we set ${\tau}$ as $0.15$.

\noindent\textbf{(b) Intensity Alignment.}
The Intensity Alignment metric evaluates the average attention intensity within the ground truth region. It is computed as:
$$
\text{Intensity Alignment} = \frac{\sum_{i, j} \mathbf{B}(i, j) \cdot \mathbf{M}(i, j)}{\sum_{i, j} \mathbf{B}(i, j)}
$$
This score reflects the degree to which the model focuses its attention on the ground truth segment, considering the intensity values of the attention map.

\begin{table}[h!]
\centering
\caption{Results on different medical image modalities.}
\label{tab:med_modality}
\vspace{-0.1in}
\resizebox{1\columnwidth}{!}{
\begin{tabular}{c|cc|cc|cc}
\toprule
\multirow{2}{*}{\textbf{Method}} & \multicolumn{2}{c|}{\textbf{Chest X-ray}} & \multicolumn{2}{c|}{\textbf{Abdomen CT}} & \multicolumn{2}{c}{\textbf{Brain MRI}} 
\\ \cline{2-7}
& \textbf{Open} & \textbf{Closed} & \textbf{Open} & \textbf{Closed} & \textbf{Open} & \textbf{Closed} \\
\midrule
\cellcolor{gray!15}LLaVA-Med + LoRA & \cellcolor{gray!15}76.96 & \cellcolor{gray!15}89.08 & \cellcolor{gray!15}79.67 & \cellcolor{gray!15}81.15 & \cellcolor{gray!15}79.15 & \cellcolor{gray!15}79.37 \\

Greedy & 79.53 & \textbf{89.92} & 81.50 & 83.61 & 80.57 & 82.54 \\
Beam & 79.83 & \textbf{89.92} & \ul{81.76} & \textbf{84.43} & 78.90 & 84.12 \\
Nucleus & 80.95 & 87.40 & 79.49 & 79.51 & 79.72 & 80.95 \\
DoLa & 78.53 & \textbf{89.92} & 81.50 & 83.61 & 79.89 & 80.95 \\
VCD & \ul{81.39} & 86.55 & 80.54 & 80.33 & 79.62 & 69.84 \\
M3ID & 79.54 & 85.71 & 76.70 & 80.33 & 80.57 & 73.01 \\
AVISC & 80.52 & 87.40 & 81.41 & 78.69 & 80.49 & \ul{85.71} \\
OPERA & 80.19 & \textbf{89.92} & 81.15 & 83.61 & 77.55 & 82.54 \\
DAMRO & 79.23 & 88.24 & 80.19 & 83.61 & 75.60 & 80.95 \\
PAI & 79.78 & \textbf{89.92} & 80.45 & 81.15 & \ul{81.95} & 80.95 \\

\ours (ours) & \textbf{82.21}& \textbf{89.92}  & \textbf{83.33} & \textbf{84.43} & \textbf{82.41} & \textbf{87.30} \\
\bottomrule
\end{tabular}}
\vspace{-0.1in}
\end{table}

\label{appd:question_types}
\begin{figure}[h!]
  \centering
  \includegraphics[width=0.7\linewidth]{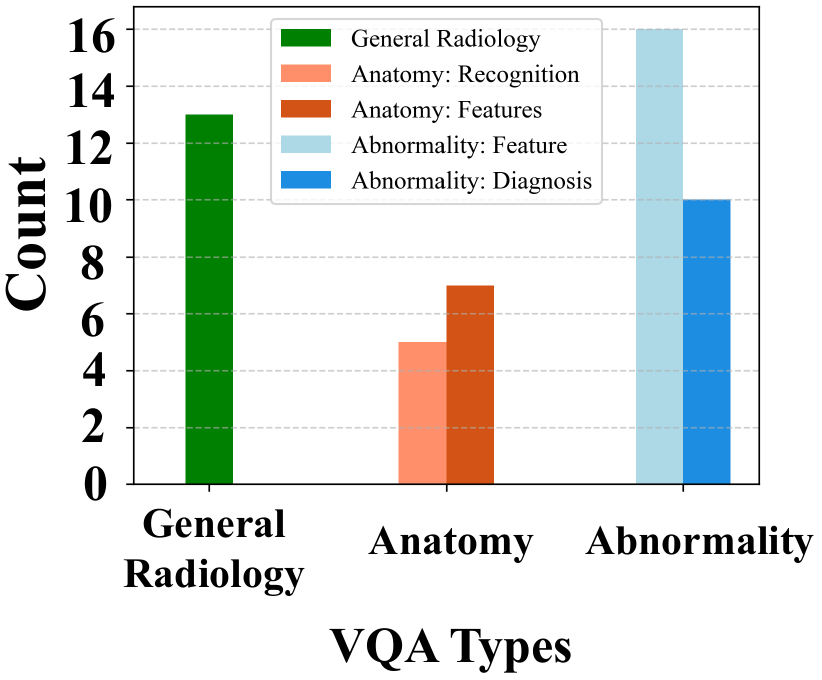}
  \vspace{-0.1in}
  \caption{VQA Types where \ours outperforms LLaVA-Med + LoRA across three image modalities.}
  \label{fig:vqa_types}
\end{figure}

\begin{figure*}[h!]
  \centering
  \includegraphics[width=0.95\linewidth]{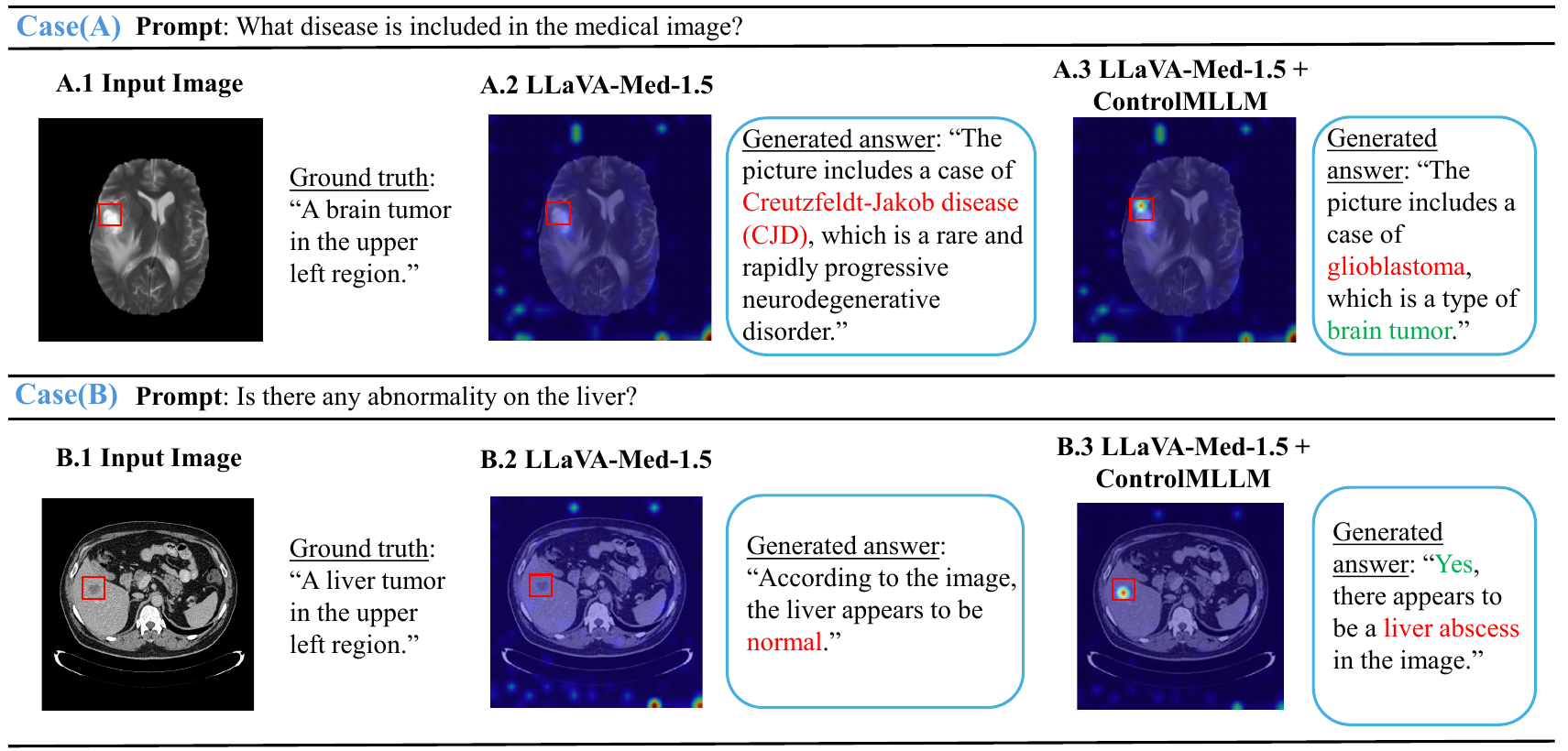}
  \vspace{-0.1in}
  \caption{Case study on LLaVA-Med-1.5. The red box in the \textbf{Input Image} (not provided as input to the model) highlights the RoI that model should focus on. Red texts and green texts indicate wrong answers and correct answers, respectively.}
  \label{fig:case_study_1.5}
  \vspace{-0.15in}
\end{figure*}

\begin{table*}[h!]
\centering
\caption{Results on report generation benchmarks, based on LLaVA-Med-1.5. }
\label{tab:report_app01_llavamed1.5}
\vspace{-0.1in}
\resizebox{1.85\columnwidth}{!}{
\begin{tabular}{c|c|ccccccc}
\toprule
\multirow{2}{*}{\textbf{Model}} & \multirow{2}{*}{\textbf{Method}} & \multicolumn{7}{c}{\textbf{IU-Xray}}\\
\cline{3-9}
 & & \textbf{BLEU} & \textbf{\text{ROUGE-L}} & \textbf{METEOR} & \textbf{BERTScore} & \textbf{CheXbert} & \textbf{RadGraph} & \textbf{RaTEScore}\\
\midrule
% \multicolumn{8}{c}{\textit{Method based on LLaVA-Med-1.5 Without Fine-tuning}} \\
% \midrule
\multirow{11}{*}{\textbf{LLaVA-Med-1.5}} & 
\cellcolor{gray!15}LLaVA-Med-1.5 & \cellcolor{gray!15}1.40 & \cellcolor{gray!15}12.41 & \cellcolor{gray!15}16.30 & \cellcolor{gray!15}84.55 & \cellcolor{gray!15}38.20 & \cellcolor{gray!15}7.77 & \cellcolor{gray!15}41.15  \\
& Greedy  & 1.04 & 12.15 & 9.87 & 85.43 & 38.04 & 5.43 & 34.97   \\
& Beam & 1.09 & 11.17 & 19.59 & 83.43 & 40.13 & 9.42 & 48.64 \\
& Nucleus & 1.44 & 12.10 & 15.60 & 81.45 & 38.04 & 6.51 & 40.41 \\
& VCD & 1.42 & 12.29 & 15.72 & 84.54 & 36.57 & 6.49 & 39.93 \\
& DoLa & 0.99 & 12.15 & 9.36 & 85.64 & 38.22 & 5.40 & 34.93  \\
& OPERA & 1.13 & 11.49 & 14.63 & 83.76 & 37.38 & 1.41 & 35.96 \\
& AVISC & 1.18 & 11.32 & 16.66 & 83.76 & 35.83 & 6.63 & 40.36  \\
& M3ID & 1.33 & 12.31 & 16.31 & 84.45 & 37.54 & 6.42 & 40.41 \\
& DAMRO & 1.27 & 11.56 & 16.42 & 84.08 & 35.60 & 6.80 & 40.08 \\
& PAI & 1.11 & 12.05 & 10.99 & 85.03 & 37.56 & 5.21 & 34.83 \\
\midrule
% \multicolumn{8}{c}{\textit{Method based on LLaVA-Med-1.5 With LoRA Fine-tuning}} \\
% \midrule
\multirow{12}{*}{\textbf{\makecell{LLaVA-Med-1.5 \\ + LoRA}}} & 
\cellcolor{gray!15}LLaVA-Med-1.5 + LoRA & \cellcolor{gray!15}8.04  & \cellcolor{gray!15}26.52 & \cellcolor{gray!15}30.37 & \cellcolor{gray!15}88.24 & \cellcolor{gray!15}51.32 & \cellcolor{gray!15}20.35 & \cellcolor{gray!15}56.97 \\
& Greedy & 9.36  & 27.57 & 27.91 & \textbf{88.55} & 52.44 & 21.28 & 58.61  \\
& Beam & \ul{9.54}  & \ul{28.41} & \ul{35.40} & 88.45 & \ul{53.70} & \ul{22.43} & \ul{59.65} \\
& Nucleus & 7.80  & 26.72 & 30.33 & 88.28 & 52.73 & 20.85 & 57.84 \\
& VCD & 8.83  & 27.36 & 31.77 & 88.30 & 51.86 & 22.02 & 58.93 \\
& DoLa & 8.93  & 26.94 & 25.74 & 88.42 & 52.27 & 20.63 & 58.10  \\
& OPERA & 9.23 & 27.48 & 34.17 & 88.17 & 51.65 & 21.37 & 57.89 \\
& AVISC & 5.57 & 21.71 & 26.84 & 87.34 & 47.32 & 16.87 & 53.66 \\
& M3ID & 8.44  & 26.21 & 30.86 & 88.20 & 51.13 & 20.77 & 59.37 \\
& DAMRO & 8.21 & 25.77 & 30.58 & 88.09 & 50.10 & 22.33 & 57.31 \\
& PAI & 8.52  & 26.97 & 28.63 & 88.42 & 52.22 & 20.99 & 58.21  \\
% \midrule
& \ours(ours) & \textbf{10.51} & \textbf{28.76} & \textbf{35.74} & \ul{88.51} & \textbf{53.88} & \textbf{23.10} & \textbf{59.66} \\
\bottomrule
\end{tabular}}
\vspace{-0.1in}
\end{table*}

\begin{table*}[h!]
\centering
\caption{Results of Report Generation on MIMIC-CXR}
\label{tab:report_app02_llavamed1.5}
\vspace{-0.1in}
\resizebox{1.85\columnwidth}{!}{
\begin{tabular}{c|c|ccccccc}
\toprule
\multirow{2}{*}{\textbf{Model}} & \multirow{2}{*}{\textbf{Method}} & \multicolumn{7}{c}{\textbf{MIMIC-CXR}}\\
\cline{3-9}
 & & \textbf{BLEU} & \textbf{\text{ROUGE-L}} & \textbf{METEOR} & \textbf{BERTScore} & \textbf{CheXbert} & \textbf{RadGraph} & \textbf{RaTEScore}\\
\midrule
% \multicolumn{8}{c}{\textit{Method based on LLaVA-Med-1.5 Without Fine-tuning}} \\
% \midrule
\multirow{11}{*}{\textbf{LLaVA-Med-1.5}} & 
\cellcolor{gray!15}LLaVA-Med-1.5 & \cellcolor{gray!15}0.98 & \cellcolor{gray!15}10.73 & \cellcolor{gray!15}10.95 & \cellcolor{gray!15}82.29 & \cellcolor{gray!15}14.11 & \cellcolor{gray!15}1.83 & \cellcolor{gray!15}31.90 \\
& Greedy  & 0.93 & 10.90 & 9.45 & 83.09 & 14.11 & 1.09 & 28.07   \\
& Beam & 1.13 & 10.95 & 13.01 & 82.34 & 12.51 & 1.87 & 32.86 \\
& Nucleus & 0.96 & 10.55 & 10.70 & 78.53 & 14.11 & 1.64 & 31.54 \\
& VCD & 1.00 & 10.93 & 11.31 & 83.11 & 13.46 & 1.97 & 31.75 \\
& DoLa & 0.65 & 9.88 & 7.83 & 83.47 & 14.07 & 1.09 & 28.07  \\
& OPERA & 1.09 & 11.51 & 12.22 & 82.69 & 13.00 & 0.52 & 27.77 \\
& AVISC & 1.16 & 11.14 & 12.30 & 82.61 & 14.11 & 1.97 & 32.20  \\
& M3ID & 1.06 & 11.50 & 11.54 & 83.15 & 13.98 & 2.24 & 32.74 \\
& DAMRO & 1.14 & 11.01 & 12.35 & 82.80 & 13.73 & 2.26 & 32.18 \\
& PAI & 1.12 & 11.67 & 10.63 & 82.80 & 14.05 & 0.98 & 28.07 \\
\midrule
% \multicolumn{8}{c}{\textit{Method based on LLaVA-Med-1.5 With LoRA Fine-tuning}} \\
% \midrule
\multirow{13}{*}{\textbf{\makecell{LLaVA-Med-1.5 \\ + LoRA}}} & 
\cellcolor{gray!15}LLaVA-Med-1.5 + LoRA & \cellcolor{gray!15}3.50  & \cellcolor{gray!15}16.38 & \cellcolor{gray!15}18.95 & \cellcolor{gray!15}85.56 & \cellcolor{gray!15}21.45 & \cellcolor{gray!15}9.54  & \cellcolor{gray!15}40.45 \\
& Greedy  & 3.50  & 16.49 & 18.71 & 85.54 & 23.43 & 9.63  & 40.49   \\
& Beam & 3.66  & 16.85 & \ul{20.68} & 85.51 & \ul{25.00} & \ul{9.91}  & \ul{41.46} \\
& Nucleus & 3.48  & 16.35 & 18.93 & 85.50 & 22.21 & 9.27 & 40.08 \\
& VCD & \ul{3.74}  & \ul{16.88} & 19.03 & 85.56 & 22.98 & 9.56  & 40.93 \\
& DoLa & 3.48  & 16.45 & 18.66 & 85.54 & 23.34 & 9.52  & 40.49  \\
& OPERA & 3.56 & 16.77 & 20.10 & 85.46 & 24.31 & 9.81 & 41.33 \\
& AVISC & 3.31  & 16.36 & 18.64 & 85.48 & 23.31 & 9.02 & 40.36 \\
& M3ID & 3.14  & 16.13 & 18.52 & 85.39 & 22.42 & 9.15 & 39.74 \\
& DAMRO & 3.42 & 16.63 & 18.87 & 85.48 & 23.30 & 9.46 & 40.80 \\
& PAI & 3.63  & 16.65 & 18.61 & \ul{85.60} & 24.51 & 9.60 & 40.49 \\
% \cline{2-9}
& \ours (ours)& \textbf{4.22} & \textbf{18.02} & \textbf{20.69} & \textbf{85.75} & \textbf{25.37} & \textbf{10.52} & \textbf{42.15} \\
\bottomrule
\end{tabular}}
% \vspace{-0.1in}
\end{table*}

\section{Full Results on Report Generation Benchmarks}
\label{appd:full_report_generation}
We present the comparison results of report generation using LLaVA-Med in Table~\ref{tab:report_app01} (IU-Xray) and Table~\ref{tab:report_app02} (MIMIC-CXR). These tables include the original results of baselines applied to LLaVA-Med without fine-tuning, where the model performs significantly worse on report generation. For example, in Table~\ref{tab:report_app01}, the best-performing baseline, M3ID, achieves a BLEU score of 1.40, which is much lower than fine-tuned LLaVA-Med baselines. This highlights the challenge of generating professional medical reports without fine-tuning. As discussed in the experiments section, \ours consistently achieves the best performance, outperforming all baselines by a large margin.

\section{Fine-grained Effectiveness Analysis}
\label{appd:finegrained_effectiveness}

In this experiment, we conduct fine-grained analyses to evaluate the effectiveness of our proposed method across different medical image modalities. Specifically, we examine how our approach improves model performance on Chest X-ray, Abdomen CT, and Brain MRI images from SLAKE. As shown in Table~\ref{tab:med_modality}, our method outperforms all baselines, particularly on Brain MRI. These
results demonstrates the effectiveness of \ours across diverse medical images and its generalization ability in medical applications.
To explain this improvement, we also include case studies with attention map visualizations for each modality in Section~\ref{sec:case_study}.

Furthermore, we analyze the VQA types where our method outperforms fine-tuned LLaVA-Med with LoRA across the three image modalities. As shown in Figure~\ref{fig:vqa_types}, our approach improves the model's performance in three key areas sensitive to image information: 
\begin{itemize}
    \item General Radiology Knowledge: Understanding medical modality types and their features.  
    \item Anatomical Structures: Recognizing features of key anatomical structures, such as organ count, location.  
    \item Abnormalities: Identifying and diagnosing abnormalities based on features like location and color. 
\end{itemize}

These improvements highlight the effectiveness of attention tuning in handling VQA types relying on accurate visual information and interaction.

\section{Analysis on LLaVA-Med-1.5}
\label{appd:exp_llavamed1.5}

\subsection{Attention Biases and Hallucination Issues in LLaVA-Med-1.5}
% \ul{Firstly include some cases to demonstrate that the attention distribution problem also exists in LLaVA-Med 1.5 and Xray-GPT} 

The attention biases we address are not unique to LLaVA-Med but are prevalent across Med-LVLMs. For instance, we include cases from LLaVA-Med-1.5 in this section. Although LLaVA-Med-1.5 is an enhanced version of LLaVA-Med, with improved training data and an increased number of visual tokens (from 256 to 576), the attention biases and hallucination issues persist.

In Figure~\ref{fig:case_study_1.5}, we visualize the attention biases and corresponding hallucination issues in LLaVA-Med-1.5. As shown, the model often fails to focus on the correct RoIs and generates hallucinated outputs. Even with attention tuning via ControlMLLM, hallucinations persist. For example, in Case (B.3), while the model identifies an abnormality in the liver, it incorrectly classifies it as a liver abscess instead of liver cancer.

These attention biases are common issues in Med-LVLMs, underscoring the need for continued research on emergent Med-LVLMs to mitigate such challenges effectively.

\subsection{Experiment Results on LLaVA-Med-1.5}

Similar to the LLaVA-Med experiments in Section~\ref{sec:report_results}, Table~\ref{tab:report_app01_llavamed1.5} and Table~\ref{tab:report_app02_llavamed1.5} show that \ours outperforms all baselines across  almost  all metrics on both IU-Xray and MIMIC-CXR, excelling in both language quality and clinical accuracy. These results underscore \ours' effectiveness in medical applications across diverse Med-LVLMs, demonstrating its strong generalization ability.

% \section{Experiment Results on XrayGPT}
% \label{appd:exp_xraygpt}

\end{document}